\documentclass[11pt,letterpaper]{article}

\usepackage[top=1in, bottom=1in, left=1in, right=1in]{geometry}
\usepackage{natbib}

\usepackage[utf8]{inputenc} 
\usepackage[T1]{fontenc}    
\usepackage[draft]{hyperref}       
\usepackage{url}            
\usepackage{booktabs}       
\usepackage{amsfonts}       
\usepackage{nicefrac}       
\usepackage{microtype}      
\usepackage{algpseudocode}

\usepackage{amsmath}
\usepackage{amsthm}
\usepackage{amssymb}
\usepackage{mathabx}
\usepackage{bm}
\usepackage{graphicx}
\usepackage{xcolor}
\usepackage[ lined]{algorithm2e}
\makeatletter
\renewcommand{\@algocf@capt@plain}{above}
\makeatother
\SetKwInput{kwInput}{Input}
\usepackage[inline]{enumitem}
\usepackage{subcaption}

\newcommand{\newnote}[1]{{\color[rgb]{0,0,0}{#1}}}
\newcommand{\mynote}[1]{{\color[rgb]{0,0,0}{#1}}}

\newcommand{\calA}{\mathcal{A}}
\newcommand{\calF}{\mathcal{F}}
\newcommand{\calB}{\mathcal{B}}
\newcommand{\bbeta}{\bm{\beta}}

\DeclareMathOperator{\dom}{dom}
\DeclareMathOperator{\mb}{MB}
\DeclareMathOperator{\child}{child}
\newcommand{\real}{\mathbb{R}}

\newcommand{\by}{\textbf{y}}

\newcommand{\bq}{\textbf{q}}
\newcommand{\bz}{\textbf{z}}

\newcommand{\bX}{\textbf{X}}

\newcommand{\bA}{\textbf{A}}

\newcommand{\supp}{\text{support}}

\newcommand{\T}{\intercal}
\newcommand{\E}{\mathbb{E}}
\newcommand{\prob}{\mathbb{P}}
\newcommand{\calP}{\mathcal{P}}
\newcommand{\calO}{\mathcal{O}}

\newtheorem{theorem}{Theorem}

\newtheorem{lemma}{Lemma}
\newtheorem{assumption}{Assumption}

\newtheorem{definition}{Definition}
\newtheorem{problem}{Problem}

\title{Learning Bayesian Networks with Low Rank Conditional Probability Tables}
\date{}

\author{%
        Adarsh Barik \\
        Department of Computer Science\\
        Purdue University\\
        West Lafayette, Indiana, USA\\
        \texttt{abarik@purdue.edu} \\
   \and
        Jean Honorio \\
        Department of Computer Science \\
        Purdue University \\
        West Lafayette, Indiana, USA\\
        \texttt{jhonorio@purdue.edu} \\
}

\begin{document}

\maketitle

\begin{abstract}
   In this paper, we provide a method to learn the directed structure of a Bayesian network using data. The data is accessed by making conditional probability queries to a black-box model. We introduce a notion of simplicity of representation of conditional probability tables for the nodes in the Bayesian network, that we call ``low rankness''. We connect this notion to the Fourier transformation of real valued set functions and propose a method which learns the exact directed structure of a `low rank` Bayesian network using very few queries. We formally prove that our method correctly recovers the true directed structure, runs in polynomial time and only needs polynomial samples with respect to the number of nodes. We also provide further improvements in efficiency if we have access to some observational data.
\end{abstract}

\section{Introduction}
\label{sec_introduction}

\paragraph{Motivation.} Real-world systems are made of large number of constituent variables. Understanding the interactions and relationships of these variables is key to understand the behavior of such systems. Scientists and researchers from many domains have been using graphs to model and learn relationships amongst variables of real-world systems for a long time. \emph{Bayesian networks} are one of the most important class of probabilistic graphical models which are used to model complex systems. They provide a compact representation of joint probability distributions among a set of variables.

\paragraph{Related work.} Learning the structure of a Bayesian network from observational data is a well known but incredibly difficult problem to solve in the machine learning community. Due to its popularity and applications, a considerable amount of work has been done in this field. Most of these work use \emph{observational data} to learn the structure. We can broadly divide these methods in two categories. The methods in the first category use score maximization techniques to learn the DAG from observational data. In this category, there are some heuristics based approaches such as \cite{friedman1999learning,tsamardinos2006max,margaritis2000bayesian,moore2003optimal} which run in polynomial-time without offering any convergence/consistency guarantee. There are also some exact but exponential time score maximizing exact algorithms such as \cite{koivisto2004exact,silander2012simple,cussens2012bayesian,jaakkola2010learning}. The methods in the second category are independence test based methods such as \cite{spirtes2000causation,cheng2002learning,yehezkel2005recursive,xie2008recursive}. 

There have also been some work to learn the structure of a Bayesian network using \emph{interventional data} \cite{murphy2001active,tong2001active,eaton2007exact,triantafillou2015constraint}. Most of these works first find a Markov equivalence class from observational data and then direct the edges using interventions. Unfortunately, the first step of finding Markov equivalence class remains NP-hard \citep{chickering1996learning}. 
\cite{hausar2012optimal}, \cite{he2008active}, \cite{kocaoglu2017experimental} have presented  polynomial time methods to find an optimal set of interventions for chordal DAGs. \cite{bello2018computationally} have proposed a method to learn a Bayesian network using interventional path queries with logarithmic sample complexity. However, their method runs in exponential time in terms of the number of parents.


In this paper, our work takes an intermediate path. We do not use pure observational or interventional data directly. Rather, we assume that there exists a \emph{black-box} which answers conditional probability queries by outputting observational data. Our goal is to limit the number of such queries and learn the directed structure of a Bayesian network. We propose a novel algorithm to achieve this goal. We also provide a method to improve our results by having access to some observational data. We intend to measure our performance based on the following criteria.
\begin{enumerate*}
	\item {\bf Correctness -} We want to come up with a method which correctly recovers the directed structure of a Bayesian network with provable theoretical guarantees.
	\item {\bf Computational efficiency -} The method must run fast enough to handle the high dimensional cases. Ideally, we want to have polynomial time complexity with respect to the number of nodes.
	\item {\bf Sample complexity -} We would like to use as few samples as possible for recovering the structure of the Bayesian network. As with the time complexity, we want to achieve polynomial sample complexity with respect to the size of the network.
\end{enumerate*}
 
\paragraph{Contribution.}
Consider a binary node $i$ of a Bayesian network with $m$ parents. The conditional probability table (CPT) of node $i$ has $2^{m + 1}$ entries. This number quickly becomes very large even for modest values of $m$. To handle such large tables while still maintaining the effect of all the parents, we introduce a notion of simplicity of representation of the CPTs, which we call ``low rankness''. Our intuition is that each CPT can be treated as summation of multiple simple tables, each of them depending only on a handful of parents (say $k$ parents where $k$ is the rank of the CPT). We connect this notion of rank of a CPT to the Fourier transformation of a specific real valued set function \citep{stobbe2012learning} and use compressed sensing techniques \citep{rauhut2010compressive} to show that the Fourier coefficients of this set function can be used to learn the structure of the Bayesian network. While doing so, we provide a method with theoretical guarantees of correctness, and which works in polynomial time and sample complexity. Our method requires computation of conditional probabilities from data. We do this by making queries to a black-box. One query consists of two steps. The first step is the selection of variables, i.e., choosing a target variable and a set of variables for conditioning. The second step is to assign specific values to the selected conditioning variables. This process is similar to the process used in \cite{bello2018computationally,kocaoglu2017experimental}, which consider a particular selection of variables as one intervention. An actual setting of the variables are considered as one experiment. For example, a selection of $k$ binary variables can be assigned $2^k$ distinct values and can be queried in $2^k$ different ways. \mynote{ Our setting is similar to an interventional setting where a selection can be compared to an intervention and an assignment can be compared to an experiment, although our method never queries the $2^k$ distinct values, but a single random assignment instead. Thus, we compare our results to the state-of-the-art interventional methods in Table~\ref{table_sample_and_time}. It should be noted that the number of queries (or experiments in the interventional setting) are a better metric for comparison than the number of selections (or interventions). This is because a selection may involve only one node \citep{bello2018computationally} or multiple nodes (in this paper) and thus could hide some complexity of the problem. Furthermore, the sample complexity of the problem depends on the number of queries.}

\begin{table}[!h]
	\caption{Sample and time complexity, number of selections (interventions) and queries (experiments) required for structure learning of binary Bayesian networks. Here $n$ is the number of nodes, $k$ is the maximum size of the Markov blanket. The maximum number of parents of a node is $\calO(k)$.}
	\label{table_sample_and_time}
	\centering
	\begin{small}
		\setlength{\tabcolsep}{2pt}
		\begin{tabular}{lllll}
			\toprule
			Algorithms & Sample Complexity & Time Complexity & Selections & Queries \\
			\midrule
			Our Work  & Blackbox - $\calO(n k^3 \log^4 n (\log k $ & $ \calO(n^4 k \sqrt{n} \log n) $& $\calO(n)$ & $ \calO(n k^3 \log^4 n ) $ \\
			(no observational data) & $+ \log \log n) )$& & & \\
            Our Work  & Observational - $\calO(n) $ & $ \calO(n^4) $& $\calO(n)$ & $ \calO(n k^3\log^4k) $ \\
            (with observational data) & Blackbox - $\calO(n k^3\log^5 k) $ & $\calO(nk^4 \sqrt{k} \log k) $ & & \\
			\cite{bello2018computationally} & Interventional - $\calO(n^2 2^k \log n)$ & $ \calO(n^2 2^k \log n) $& $\calO(n^2)$ & $ \calO(n^2 2^k) $ \\
			\cite{kocaoglu2017experimental} & Interventional - no guarantees & $ \calO(2^n k n^2 \log^2 n) $ & $\calO(\log n)$ & $ \calO(2^n \log n) $ \\
			\bottomrule
		\end{tabular}
	\end{small}
\end{table}


\section{Preliminaries}
\label{sec_preliminaries}

In this section, we introduce formal definitions and notations. Let $\bX = \{X_1, X_2,\dots, X_n\}$ be a set of random variables. For a set $A$, $X_A$ denotes the set of random variables $X_i \in \bX$ such that $i \in A$. We use the shorthand notation $\overline  i$ to denote $V \setminus \{i\}$. We define a Bayesian network on a directed acyclic graph $G=(V, E)$ where $V$ denotes the set of vertices and $E$ is a set of ordered pair of nodes, each corresponding to a directed edge, i.e., if $(a, b) \in E$ then there is an edge $a \to b$ in $G$. The parents of a node $i, \forall i \in V$ denoted by $\pi_G(i)$, are set of all nodes $j$ such that edge $(j, i) \in E$. We also define the Markov blanket $\mb_G(i)$ for a node $i$ as a set of nodes containing parents, children and parents of children of node $i$. The nodes with no children are called terminal nodes.  

\begin{definition}[Bayesian network]
    \label{def_bayesian_network}
    Let $G=(V, E)$ be a directed acyclic graph (DAG) and $\bX = \{X_1, X_2,\dots, X_n\}$ be a set of random variables such that $X_i$ corresponds to a random variable at node $i \in V, \forall i=\{1,\dots,n\}$. Let $X_{\pi_G(i)}$ denote the set of random variables defined on the parents of node $i$ in DAG $G$. A Bayesian network $\calB = (G, \calP)$ represents a joint probability distribution $\calP$ over the set of random variables $\bX$ defined on the nodes of DAG $G$ which factorizes according to the DAG structure, i.e., $\calP(X_1, X_2,\dots,X_n) = \prod_{i=1}^n \calP(X_i | X_{\pi_G(i)})$
    where $\calP(X_i | X_{\pi_G(i)})$ denotes conditional probability distribution (CPD) of node $i$ given its parents in DAG $G$.
\end{definition}
We denote the domain of a random variable $X_i,  \forall i \in \{1,\dots,n\}$ by $\dom(X_i)$. The cardinality of a set is denoted by notation $|\cdot|$. A Bayesian network $\calB = (G, \calP)$ on discrete nodes is called a \emph{binary Bayesian network} if $|\dom(X_i)| = 2, \forall i \in \{1,\dots,n\}$. For discrete nodes, $\calP(X_i | X_{\pi_G(i)})$ is often represented as a conditional probability table (CPT) with $|\dom(X_i)| \prod_{j \in \pi_G(i)} |\dom(X_j)|$ entries. In this work, we will only focus on binary Bayesian networks. Next, we introduce a novel concept of rank of a conditional probability distribution for a node of Bayesian network. 

\begin{definition}[Rank $k$ conditional probability distribution]
    \label{def_rank_k_probability_distribution}
    A node $i\in V$ of a Bayesian network $\calB(G, \calP)$ is said to be rank $k$ representable with respect to a set $A(i) \subseteq V \setminus \{i\}$ and probability distribution $\calP$ if,
    \begin{align}
        \label{eq_rank_k_probability_distribution}
        \begin{split}
            \calP(X_i = x_i | X_{A(i)} = x_{A(i)}) = & \sum_{\substack{S \subseteq \{i\} \cup A(i)\\ 1 \leq |S| \leq k,\ i \in S }} Q_{S}(X_S = x_S), \forall x_i \in \dom(X_i), x_{A(i)} \in \dom(X_{A(i)})
        \end{split}
    \end{align}
    where $Q_S: \bigtimes_{j \in S} \dom(X_j) \rightarrow \real $ is a function which depends only on the variables $X_S$. A node $i$ is said to have rank $k$ conditional probability table if it is rank $k$ representable but is not rank $k-1$ representable with respect to $A(i)$ and $\calP$.
\end{definition}
For example, a node $i \in V$ of a Bayesian network $\calB(G, \calP)$ is rank $2$ representable with respect to its parents $\pi_G(i)$ and $\calP$ if we can write $\calP(X_i = x_i | X_{\pi_G(i)} = x_{\pi_G(i)}) =  Q_i(X_i = x_i) + \sum_{j \in \pi_G(i)} Q_{ij}(X_i = x_i, X_j = x_j), $
where  $\forall x_i \in \dom(X_i), x_j \in \dom(X_j), \forall j \in \pi_G(i)$. It is easy to observe that any node $i \in V$ is always rank $|A(i)| + 1$ representable with respect to a set $A(i) \subset V \setminus \{i\}$ and $\calP$. Also, rank $k$ representations for a node $i$ with respect to $A(i)$ and $\calP$ may not be unique. We consider real-valued set functions on a set $T$ of cardinality $t$ defined as $f:2^T \rightarrow \real$ where $2^T$ denotes the power set of $T$. Let $\calF$ be the space of all such functions, with corresponding inner product $\langle f, g \rangle \triangleq 2^{-t} \sum_{A \in 2^T} f(A) g(A)$. The space $\calF$ has a natural Fourier basis, and in our set function notation the corresponding Fourier basis vectors are $	\psi_B(A) \triangleq (-1)^{|A \cap B|}$.
We define the Fourier transformation coefficients of function $f$ as $	\hat f(B) \triangleq \langle f, \psi_B \rangle = 2^{-t} \sum_{A \in 2^T} f(A) (-1)^{|A \cap B|}$.
Using Fourier coefficients, the function $f$ can be reconstructed as: 
\begin{align}
	\label{eq_fourier_reconstruction}
	f(A) = \sum_{B \in 2^T} \hat f(B) \psi_B(A)
\end{align}
The Fourier support of a set function is the collection of subsets with nonzero Fourier coefficient: $\supp(\hat f) \triangleq \{B \in 2^T | \hat f(B) \ne 0 \}$.

\section{Method and Theoretical Analysis}
\label{sec_method_and_theoretical_analysis}

In this section, we develop our method for learning the structure of a Bayesian network and provide theoretical guarantees for correct and efficient learning. First we would like to mention some technical assumptions.
%
%
\begin{assumption}[Availability of Black-box]
	\label{assum_intervention}
	For a Bayesian network $\calB(G, \calP)$, we can submit a conditional probability query $BB(i,A,x_A,N)$ to a black-box on any set of selected nodes $i \in V, A \subseteq \overline {i}$ and value $x_A$, and receive $N$ i.i.d. samples from the conditional distribution $\calP(X_i|X_A=x_A)$.
\end{assumption}
\begin{assumption}[Faithfulness]
	\label{assum_faithfulness}
	The distribution over the nodes of the Bayesian network $\calB(G, \calP)$ induced by $(G, \calP)$ exhibits no other independencies beyond those implied by the structure of $G$. 
\end{assumption}
\begin{assumption}[Low rank CPTs]
	\label{assum_low_rank}
	Each node $i \in V$ in the Bayesian network $\calB(G, \calP)$ has rank $2$ conditional probability tables with respect to $\pi_G(i)$ and $\calP$.  
\end{assumption}

\mynote{Assumption \ref{assum_intervention} implies the availability of observational data for all queries. This is analogous to the standard assumption of availability of interventional data in interventional setting \citep{murphy2001active,he2008active,kocaoglu2017experimental,tong2001active,hausar2012optimal}.} Assumption \ref{assum_faithfulness} is also a standard assumption \citep{kocaoglu2017experimental,tong2001active,he2008active,spirtes2000causation,triantafillou2015constraint} which ensures that we only have those independence relations between nodes which come from d-separation. We also introduce a novel Assumption \ref{assum_low_rank} which ensure that CPTs of nodes have a simple representation. In the later sections, we relate this to sparsity in the Fourier domain. We note that there is nothing special about CPTs being rank $2$ and our method can be extended for any rank $k$ CPTs.

\subsection{Problem Description}
\label{subsec_problem}

In this work, we address the following question:
\begin{problem}[Recovering structure of a Bayesian network using black-box queries]
	Consider we have access to a black-box which provides observational data for our conditional probability queries for a faithful Bayesian network $\calB(G, \calP)$ with each node $i$ having rank $2$ CPT with respect to its parents $\pi_G(i)$ and $\calP$. Can we recover the directed structure of $G$ with theoretical guarantees of correctness and efficiency in terms of time and sample complexity? 
\end{problem}
We show that it is indeed possible to do. We control the number of samples by controlling the number of queries. We also show that it is possible to further reduce the sample complexity if we have access to some observational data.

\subsection{Theoretical Result}
\label{subsec_theoretical_result}

In this subsection, we state our theoretical results. We start by analyzing terminal nodes.

\paragraph{Analyzing Terminal Nodes.} 
Since terminal nodes do not have any children, their Markov blanket only contains their parents. Furthermore, if the Bayesian network is faithful then for any terminal node $t \in V$: $\calP(X_t | X_{\pi_G(t)}) = \calP(X_t | X_{\mb_G(t)}) = \calP(X_t | X_{\overline  t})$.
Thus, for any terminal node $t \in V,\ \calP(X_t | X_{\pi_G(t)})$ can be computed without explicitly knowing its parents. Next, we define a set function which computes $\calP(X_t | X_{\pi_G(t)})$. In particular, for an assignment $x_{\pi_G(t)} \in \{0, 1\}^{|\pi_G(t)|}$, we are interested in computing $\calP(X_t = 1 | X_{\pi_G(t)} = x_{\pi_G(t)})$. Note that  $\calP(X_t = 0 | X_{\pi_G(t)} = x_{\pi_G(t)})$ can simply be computed by subtracting $\calP(X_t = 1 | X_{\pi_G(t)} = x_{\pi_G(t)})$ from $1$. Let $\overline t$ denote the set $V \setminus \{t\}$. For node $t$ and a set $A \subseteq \overline t$, let $x^A \in \{0,1\}^n$ be an assignment such that $x_i^A = \bm{1}_{i \in A}, \forall i \ne t$ and $x_t^A = 0$.
We define a set function $f_t$ for each terminal node $t \in V$ as follows:
\begin{align}
    \label{eq_set_function_terminal_nodes}
    f_t(A) = Q_t(X_t = x_t^A) + \sum_{j \in \pi_G(t)} Q_{tj}(X_t = x_t^A, X_j = x_j^A), \quad \forall A \subseteq \overline t
\end{align}
Note that Equation \eqref{eq_set_function_terminal_nodes} precisely computes $\calP(X_t = 1 | X_{\pi_G(t)} = x_{\pi_G(t)}^A)$ and $f_t(A) = f_t(A \cap \pi_t)$.
Next, we prove that the Fourier support of $f_t$ only contains singleton sets of parents of node $t$.

\begin{theorem}
    \label{thm_terminal_nodes_with_rank_2}
    If nodes of a Bayesian network $\calB(G, \calP)$ have rank $2$ with respect to their parents $\pi_G(.)$ and $\calP$, then the Fourier coefficient $\hat f_t(B)$ for function $f_t$ defined by equation \eqref{eq_set_function_terminal_nodes} for any terminal node $t$ and a set $B \in 2^{\overline t}$ is given by:
    \begin{align}
        \label{eq_fourier_coeff_terminal_nodes}
        \hat f_t(B) = \begin{cases}
                            Q_t(X_t = 1) + \frac{1}{2} \sum_{j \in \pi_G(t)} \big( Q_{tj}(X_t = 1, X_j = 0) + Q_{tj}(X_t = 1, X_j = 1)  \big), \quad B = \phi \\
                            \frac{1}{2} \big( Q_{tj}(X_t = 1, X_j = 0) - Q_{tj}(X_t = 1, X_j = 1) \big), \quad B =\{j\}, \forall j \in \pi_G(t) \\
                            0, \quad \text{Otherwise}
                      \end{cases}
    \end{align}
\end{theorem}
(See Appendix \ref{proof_detailed_proofs} for detailed proofs.)

\paragraph{Analyzing Non-Terminal Nodes.}

A similar analysis can be done for non-terminal nodes. However, for a non-terminal node $i$ we can not compute $\calP(X_i | X_{\pi_G(i)})$ without explicitly knowing the parents of node $i$. We will rather focus on computing $\calP(X_i | X_{\mb_G(i)})$ for non-terminal 
nodes which equals to computing $\calP(X_i | X_{\overline  i})$ which can be done from data. Similar to the previous case, we define a set function $g_i$ for each non-terminal node $i \in V$ as follows:
\begin{align}
g_i(A) = Q_i(X_i = x_i^A) + \sum_{j \in \pi_G(i)} Q_{ij}(X_i = x_i^A, X_j = x_j^A), \quad \forall A \subseteq \overline i
\end{align} 
We can define a corresponding set function $f_i$ which computes $\calP(X_i|X_{\mb_G(i)})$ for non-terminal nodes. We do it in the following way: 
\begin{align}
\label{eq_set_function_nonterminal_nodes}
\begin{split}
f_i(A) = & \calP(X_i = x_i^A |X_{\mb_G(i)} = x_{\mb_G(i)}) \\
= & \frac{\calP(X_i = x_i^A | X_{\pi_G(i)} = x_{\pi_G(i)}^A) \prod_{k \in \child_G(i)} \calP(X_k = x_k^A | X_{\pi_G(k)} = x_{\pi_G(k)}^A)}{\sum_{X_i} \calP(X_i = x_i^A| X_{\pi_G(i)} = x_{\pi_G(i)}^A) \prod_{k \in \child_G(i)} \calP(X_k = x_k^A | X_{\pi_G(k)} = x_{\pi_G(k)^A} )} 
\end{split}
\end{align}
where $\child_G(i)$ is the set of children of node $i$ in DAG $G$. We can again compute the Fourier support for $f_i$ for each non-terminal node.

\begin{theorem}
	\label{thm_nonterminal_nodes_with_rank_2}
	If nodes of a Bayesian network $\calB(G, \calP)$ have rank $2$ with respect to their parents $\pi_G(.)$ and $\calP$, then the Fourier coefficient $\hat f_i(B)$ for function $f_i$ defined by equation \eqref{eq_set_function_nonterminal_nodes} for any non-terminal node $i$ and a set $B \in 2^{\overline i}$ is given by:
	\begin{align}
	\label{eq_fourier_coeff_nonterminal_nodes}
	\hat f_i(B) = \begin{cases}
	0, \quad | B\setminus \mb_G(i) | \geq 1  \\
	\frac{1}{2^{n - 1}} \sum_{A \in 2^{V - i}}  \frac{ g_i(A ) {\prod_{k \in \child_G(i)}} g_k(A) }{g_i(A ) {\prod_{k \in \child_G(i)}} g_k(A) +  g_i(A \cup \{i\}) {\prod_{k \in \child_G(i)}} g_k(A \cup \{i\}) } \psi_B(A) , \quad \text{otherwise}
	\end{cases}
	\end{align}
\end{theorem} 

\subsection{Algorithm}
\label{subsec_algorithms}

Our algorithm works on the principle that the terminal nodes are rank $2$ with respect to their Markov Blanket and $\calP$, while non-terminal nodes are not. This is true if for every non-terminal node there exists a $B \in 2^{V}$ such that $|B \setminus \mb_G(i)| = 0$ and $\hat f_i(B)$ is nonzero. This is formalized in what follows.
\begin{assumption}[Non-terminal nodes are not rank $2$]
	\label{assum_non_terminal_not_rank_2}
	There exists a $B \in 2^V$ for each non-terminal node $i$, such that $|B| = 2,\ |B \setminus \mb_G(i)| = 0$ and $\hat f_i(B)$ as defined by Equation \eqref{eq_fourier_coeff_nonterminal_nodes} is non-zero.
\end{assumption}

This distinction helps us to differentiate between terminal and non-terminal nodes. Note that the set function $f_i$ is uniquely determined by its Fourier coefficients. Moreover, the Fourier support for function $f_i$ is sparse. For terminal nodes, $\hat f_i(B)$ is non-zero only for the empty set or the singleton nodes, while for the non-terminal nodes, $\hat f_i(B)$ is non-zero for $B \subseteq \mb_G(i)$. Thus, recovering Fourier coefficients from the measurements of $f_i$ can be treated as recovering a sparse vector in $\real^{2^{\overline i}}$. However, $|2^{\overline i}|$ could be quite large. We avoid this problem by substituting $f_i$ by another function $g_i \in \mathcal{G}_2$ where $\mathcal{G}_k = \{ g_i \mid \forall B \in \supp(g_i), |B| \leq k \}$. Note that,
\begin{align}
	\label{eq_f_i_sum_factors}
	f_i(A_j) = \sum_{\substack{|B_k|=1 \\ B_k \in 2^{\overline i}}} \hat f_i(B_k) \psi_{B_k}(A_j) + \sum_{\substack{|B_k|=2 \\ B_k \in 2^{\overline i}}} \hat f_i(B_k) \psi_{B_k}(A_j) + \sum_{\substack{|B_k|\geq 3 \\ B_k \in 2^{\overline i}}} \hat f_i(B_k) \psi_{B_k}(A_j)
\end{align}
and $\forall g_i \in \mathcal{G}_2$,
\begin{align}
\label{eq_g_i_sum_factors}
g_i(A_j) = \sum_{\substack{|B_k|=1 \\ B_k \in 2^{\overline i}}} \hat f_i(B_k) \psi_{B_k}(A_j) + \sum_{\substack{|B_k|=2 \\ B_k \in 2^{\overline i}}} \hat f_i(B_k) \psi_{B_k}(A_j)
\end{align}

%

It follows that for a terminal node $i$, $g_i = f_i$ as for terminal nodes $f_i \in \mathcal{G}_1$. \mynote{For non-terminal nodes, using results from Theorem \ref{thm_nonterminal_nodes_with_rank_2}, if $B \subseteq \mb_G(i)$ then $\hat f_i(B) \ne 0$ and therefore $g_i \notin \mathcal{G}_1$.}
Now, let $\calA_i$ be a collection of $m_i$ sets $A_j \in 2^{V - i}$ chosen unifromly at random. We measure $g_i(A_j)$ for each $A_j \in \calA_i$ and then using equation \eqref{eq_fourier_reconstruction} we can write:
\begin{align}
	g_i(A_j) = \sum_{B_k \in 2^{\overline i}, |B_k| \leq 2 } (-1)^{|A_j \cap B_k|} \hat f_i (B_k), \forall A_j \in \calA_i
\end{align}
Let $\bm{g_i} \in \real^{m_i}$ be a vector whose $j$\textsuperscript{th} row is $g_i(A_j)$ and $\bm{\hat g_i} \in \real^{n + {n-1 \choose 2}}$ be a vector with elements of form $\hat f_i(B_k) \forall B_k \in \rho_i $ where 
\begin{align}
\label{eq_rho}
\rho_i = \{B_k \mid B_k \in 2^{\overline i}, |B_k| \leq 2\}
\end{align}
is a set which contains $\supp(\hat f_i)$. Then,
\begin{align}
    \label{eq_compressed_sensing_formulation}
    \bm{g_i} = \mathcal{M}_i \bm{\hat g_i}\quad \text{ where, } \mathcal{M}_i \in \{-1, 1\}^{m_i \times n} \text{ such that } \mathcal{M}^i_{jk} = (-1)^{|A_j \cap B_k|} \ .
\end{align}
Also note that for terminal nodes $\bm{\hat g_i}$ is sparse with  $|\pi_G(i)| + 1 $ non-zero elements for terminal nodes and at max ${k \choose 2} + k + 1$ non-zero elements for non-terminal nodes where $k = |\mb_G(i)|$. Equation \eqref{eq_compressed_sensing_formulation} can be solved by any standard compressed sensing techniques to recover the parents of the terminal nodes. Using this formulation and the fact that terminal nodes have non-zero Fourier coefficients on empty or singleton sets, we can provide an algorithm to identify the terminal nodes and their corresponding parents. We can use this algorithm repeatedly to identify the complete structure of the Bayesian network until the last two nodes where we can not apply our algorithm. Algorithm \ref{algo_getParents} identifies the parents for each node and consequently the directed structure of the Bayesian network.

\begin{minipage}{0.46\textwidth}
	\vspace{-7\baselineskip}
	\begin{algorithm}[H]
		\caption{\label{algo_getParents}getParents$(V)$}
		\SetKwInOut{Input}{Input}
		\SetKwInOut{Output}{Output}
		\Input{Nodes $V = \{1, 2,\dots, n\}$}
		\Output{Recovered parent set $\hat\pi: V \to 2^V$}
		$S \leftarrow V$ \;
		\While{$|S| \ge 3$}{
			$T, \hat\pi = \text{getTerminalNodes}(S)$  \;
			$S \leftarrow S \setminus T$ \;
		}
		\For{$i \in S$}{
			$\hat\pi(i) = \phi $;
			}
	\end{algorithm}
\end{minipage}%
\hfill
\begin{minipage}{0.50\textwidth}
\begin{algorithm}[H]
	\caption{\label{algo_getTerminalNodes}getTerminalNodes$(S)$}
	\SetKwInOut{Input}{Input}
	\SetKwInOut{Output}{Output}
	\Input{Nodes $S \subseteq \{1, 2,\dots, n\}$ }
	\Output{Set of terminal nodes $T$ and their parents $\hat\pi: T \rightarrow 2^{S}$}
	$T \leftarrow  \phi, \hat\pi(i) \leftarrow \phi \ \forall i \in S  $ \;
	\For{node $i \in S$, $j \in \{ 1, \dots, m_i \}$}{
			Choose $A_j \in 2^{S \setminus \{i\}}$ uniformly at random \;
			Compute $f_i(A_j) = \calP(X_i = 0| X_{S \setminus \{i\}} = x_{S \setminus \{i\}}^{A_j})$ \;
			Compute $\mathcal{M}^i_{jk}$ for  $B_k \in \rho_i$ (Eq \eqref{eq_rho} \eqref{eq_compressed_sensing_formulation}) \;
		Solve for $\bbeta_i$ using compressed sensing (Eq \eqref{eq:opt prob})  \;
		\If{$\bbeta_i(B) = 0$ for all $|B| > 1$}{
			$T \leftarrow T \cup \{i\}$ \;
			$\hat\pi(i) \leftarrow \cup_{B:  \bbeta_i(B) \ne 0} B$ \;
		}
	}
\end{algorithm}%
\end{minipage}

\section{Analysis in Finite Sample Regime}
\label{sec_analysis_in_sample_regime}

So far our results have been in the population setting where we assumed that we had access to the true conditional probabilities. However, generally this is not the case and we have to work with a finite number of samples from the \emph{black-box}. In this section, we provide theoretical results for different finite sample regimes.

\subsection{Without access to any observational data}
\label{subsec_pure_interventional_setting}

In this setting, we assume that we only have access to a black-box which outputs observational data for our conditional probability queries. One selection of nodes consists of fixing $X_{\overline  i}$ and then measuring $X_i$ for each node $i$. We need only $1$ selection for each node. Thus the total number of selections for all the nodes is $n$. One query amounts to fixing $X_{\overline  i}$ to a particular $x_{\overline  i}$. Note that while $2^{n-1}$ such queries are possible for each selection on each node, we only conduct $m_i$ queries for each node $i$.

\paragraph{Number of Queries.}
\label{subsubsec_theoretical_result}
We measure $g_i(A_j)$ by querying for $f_i(A_j)$. Let $|f_i(A_j) - g_i(A_j)| \leq \epsilon_j,  \forall A_j \in \calA_i$ for some $\epsilon_j > 0$. Once we have the noisy measurements of $g_i(A_j)$, we can get a good approximation of
$\bm{\hat g_i}$ by solving the following optimization problem for each node $i$.
\begin{align}
\label{eq:opt prob}
\bbeta_i = \min_{\bm{\hat g_i} \in \real^{|\rho_i|}} \| \bm{\hat g_i} \|_1 \quad s.t. \| \mathcal{M}_i \bm{\hat g_i} - \bm{f_i} \|_2 \leq \epsilon \quad \text{where }  \epsilon = \sqrt{\sum_{A_j \in \calA_j} \epsilon_j^2} \ .
\end{align}

\begin{theorem}
	\label{thm_Stobbe}
	Suppose $\bm{\hat g_i}$ is constructed by computing $\hat g_i(B_k)$ using $B_k$ from a fixed collection $ \rho_i$ as defined in Equation \eqref{eq_rho}. Furthermore, suppose $\bm{g_i}$ is computed by selecting $m_i$ sets $A_j$ uniformly at random from $2^{\overline i}$. We define the matrix $\mathcal{M}_i$ as in equation \eqref{eq_compressed_sensing_formulation}. Then there exist universal constants $C_1, C_2 > 0$ such that if, $m_i \geq \max( C_1 |\supp(\hat g_i)| \log^4 ( n + {{n - 1} \choose 2} ), C_2 |\supp(\hat g_i)| \log \frac{1}{\delta}  )$
	and $\bm{\beta_i}$ is solved using equation \eqref{eq:opt prob}. Then with probability at least $1 - \delta$, we have $\| \bm{\beta_i} - \bm {\hat g_i} \|_2 \leq C_3 \frac{\epsilon}{\sqrt{m_i}}$ for some universal constant $C_3 > 0$.  If the minimum non-zero element of $ |\bm{\hat g_i}| $ is greater than $2 C_3  \frac{\epsilon}{\sqrt{m_i}} $ then $\bbeta_i$ recovers $\bm{\hat g_i}$ up to the signs. Furthermore, if Assumption \ref{assum_non_terminal_not_rank_2} is satisfied then $|\bbeta_i(B)| \leq C_3  \frac{\epsilon}{\sqrt{m_i}} , \forall B \in \rho_i, |B| = 2$ if and only if $i$ is a terminal node and  $\hat{\pi}(i) = \{B \mid |B| = 1, \ |\bbeta_i(B)| > C_3  \frac{\epsilon}{\sqrt{m_i}} \} $ correctly recovers the parents of the terminal node $i$, i.e., $\hat{\pi}(i) = \pi_G(i)$. Applying this recursively shows the correctness of Algorithm \ref{algo_getParents}. 
\end{theorem}

The sparsity for each node would be less than or equal to ${k \choose 2} + k + 1$. Thus the number of queries needed for each node (using arguments from Theorem \ref{thm_Stobbe}) would be of order $\calO(\max (k^2 \log^4 n, k^2 \log \frac{1}{\delta} ) )$. At the first iteration, we query all the nodes. From the next iteration onwards, we query for only the nodes which had terminal nodes as their children,i.e., for a maximum of $k$ nodes. Thus the total number of queries needed would be $\calO( \max (n k^3 \log^4 n, n k^3 \log \frac{1}{\delta} ))$. We can recover parents for terminal nodes using Theorem \ref{thm_Stobbe}.
%
%

\paragraph{Sample and Time Complexity.}
The sample complexity is  $\calO(\max ( \frac{n k^3 \log^4 n}{\epsilon^2} (\log k + \log \log n), \frac{n k^3}{\epsilon^2}\log \frac{1}{\delta} (\log k + \log \log n  ))$  and the time complexity is $\calO(n^4 k \sqrt{n} \log n)$ (See Appendix \ref{proof_sample_time_no_obs_data} for details).

\subsection{With Access to Some Observational Data}
\label{subsec_intervention_with_access_to_some_observational_data}

In this setting, we have access to some observational data as well. We can use the observational data to figure out the Markov blanket of each node which helps us reduce number of selected variables in the conditional probability queries. Once we have the Markov blanket, we only select the nodes in $\mb_G(i)$ for each query. We need only 1 selection for each node. Thus the total number of selections for all nodes does not exceed $n$.

\paragraph{Using Observational Data.}
\label{subsubsec_observational_data}
Recall that $\calP$ is the true joint distribution over the nodes of a Bayesian network $\calB(G, \calP)$. We define a collection of distributions $\prob$ over the nodes of the Bayesian network as: $\prob = \big\{ \text{$P$ is faithful to $G$.}\ | P(X_i | X_l) = \calP(X_i | X_l), \forall i, l \in \{1, \dots, n\}  \big\}$

\paragraph{Computing the Markov Blanket from Observational Data.}
Consider a probability distribution $\hat{P} \in \prob$ on the nodes of the Bayesian network such that each node $i$ is rank $2$ with respect to $\mb_G(i)$ and $\hat{P}$. This allows us to recover the Markov blanket of the node using the observational data.

\begin{theorem}
	\label{thm_recover_mb}
	If there exists a probability distribution $\hat{P} \in \prob$ such that each node $i$ is rank $2$ with respect to $\mb_G(i)$ and $\hat{P}$, then the Markov blanket of a node $i$ can be recovered by solving the following system of equations:
	\begin{align*}
	\begin{split}
	&\calP(X_i = 0, X_l=0) =  \tilde{Q}_i(X_i = 0) \calP(X_l=0) + \sum_{\substack{j \in -i \\ j \ne l}} \tilde{Q}_{ij}(X_i = 0, X_j = 0)  \calP( X_j = 0, X_l=0) \\
	&\hspace{1.3in}+ \tilde{Q}_{il}(X_i = 0, X_l=0) \calP(X_l = 0 ),\ \forall\ l=\{1,\dots,n \}, l \ne i \\
	&\calP(X_i = 0) =  \tilde{Q}_i(X_i = 0)  + \sum_{\substack{j \in -i \\ j \ne l}} \tilde{Q}_{ij}(X_i = 0, X_j = 0)  \calP( X_j = 0)
	\end{split}
	\end{align*}
	which can be written in a more compact form:
	\begin{align}
	\label{eq_compute_Q_tilde_matrix_compact}
	\overline \by &= \overline \bA \bq 
	\end{align}
	where $\overline \by \in \real^n$ and $\overline \bA \in \real^{n \times n}$ and $\bq \in \real^n$.
\end{theorem}

 For terminal nodes, existence of $\hat P \in \prob$ as $\calP \in \prob$ is guaranteed. To ensure that $\hat P \in \prob$ also exists for non-terminal nodes, we make the following assumption:
\begin{assumption}
	\label{assum_rank_2_phat}
	The population matrix $\overline \bA \in \real^{n \times n}$ as defined in equation \eqref{eq_compute_Q_tilde_matrix_compact} is positive definite. 
\end{assumption}

This assumption is not strong. We can, in fact, show that $\overline \bA$ is a positive semidefinite matrix.

\begin{lemma}
	\label{lem_pso_sem_def_bar_A}
	The population matrix $\overline \bA$ as defined in equation \eqref{eq_compute_Q_tilde_matrix_compact} is a positive semidefinite matrix.
\end{lemma}

We can solve Equation \eqref{eq_compute_Q_tilde_matrix_compact} to get  $\tilde{Q}_i$ and $\tilde{Q}_{ij}$. The Markov blanket of node $i$ is computed by $\mb_G(i) = \{ j |\  \tilde{Q}_{ij} \ne 0 \}$. To this end, we prove that:
\begin{lemma}
	\label{lem_only_recover_mb}
	If $\tilde{Q}_{ij}(\cdot,\cdot), \forall j \in \{1,\dots,n\}, j \ne i$ is computed by solving system of linear equations \eqref{eq_compute_Q_tilde_matrix_compact} and $\hat P \in \prob$ is faithful to $G$ then $ \tilde{Q}_{ij}(\cdot,\cdot) \ne 0, \forall j \in \{1,\dots, n\}, j \ne i$ if and only if $j \in \mb_G(i)$. 
\end{lemma}
Once we know the Markov blanket for each node $i$, the queries in Algorithm \ref{algo_getTerminalNodes} can be changed from $\bm{f_i}(A_j) = \calP(X_i = 0| X_{S \setminus \{i\}} = x_{S \setminus \{i\}}^{A_j})$ to $\bm{f_i}(A_j) = \calP(X_i = 0| X_{S \cap \mb_i} = x_{S \cap \mb_i}^{A_j})$ which helps in reducing the sample and time complexity.

\paragraph{Number of Queries.}
Again, let $|\mb_G(i)| \leq k, \forall i \in \{1,\dots,n\}$. The sparsity for each node would be less than or equal to ${k \choose 2} + k + 1$. Thus number of queries needed for each node (using arguments from Theorem \ref{thm_Stobbe}) would be of order $\calO(\max(k^2 \log^4 k, k^2 \log \frac{1}{\delta})$. As before, these queries are repeated ${nk}$ times. Thus the total number of queries needed would be $\calO(\max(n k^3 \log^4 k, n k^3 \log \frac{1}{\delta} ))$.

\paragraph{Sample and Time Complexity.}
\label{subsubsec_sample_complexity}

We use the following lemma to get the sample complexity for the observational data. 
\begin{lemma}
	\label{lem_observation_samples}
	 $N = \calO(\frac{\log n}{\epsilon^2})$ i.i.d observations are sufficient to measure elements of $\overline \bA$ and $\overline \by$, $\epsilon$ close to their true value. That is $| \overline \bA - \hat{\bA} |  \leq \epsilon$ and $|\overline \by-\hat\by| \leq \epsilon$, for some $\epsilon > 0$ with probability at least $1 - 2 \exp( \log ( {n \choose 2} + 3n )  - \frac{N \epsilon^2}{2})$ for some $\epsilon > 0$ where $\hat{\bA}$ and $\hat{\by}$ are the empirical measurements of $\overline {\bA}$ and $\overline {\by}$ respectively and $|\cdot - \cdot|$ denotes componentwise comparison for matrices.
\end{lemma}

At this point, it remains to be shown that we can still recover the Markov blanket for the nodes using the noisy measurements of unary and pairwise marginals. Below, we prove that this is true as long as $\overline \bA$ is well conditioned. 

\begin{lemma}
	\label{lem_observ_nosiy_measurement}
	Let $\hat\bA$ and $\hat\by$ be the empirical measurements of $\overline \bA$ and $\overline \by$ as defined in equation \eqref{eq_compute_Q_tilde_matrix_compact} respectively such that  $|\hat\bA - \overline \bA| \leq \epsilon$ and $|\hat\by - \overline \by| \leq \epsilon$ for some $\epsilon > 0$, where $|\cdot - \cdot|$ denotes componentwise comparison for matrices. Let $\hat\bq$ be the solution to the system of linear equations given by $\hat\by = \hat\bA \hat\bq $ and $\eta \kappa_{\infty}(\overline \bA) \leq 1$, then $\hat\bq$ recovers $\bq$ up to signs as long as $N = \calO(n)$ i.i.d. measurements are used to measure $\hat{\bA}$ and $ \frac{\max_i |\bq_i|}{\min_i |\bq_i|} \leq \frac{1 - \eta \kappa_{\infty}(\overline \bA)}{4 \eta \kappa_{\infty}(\overline \bA)}$, where $\kappa_{\infty}(\overline \bA) \triangleq \|\overline \bA\|_{\infty} \| \overline \bA^{-1} \|_{\infty}$ is the condition number of $\overline \bA$ and $\eta = \max ( \frac{n \epsilon}{ \sum_{j=1}^{n-1} \calP(X_j = 0 ) + 1 }, \frac{\epsilon}{\calP(X_n = 0)})$.
\end{lemma}

The time complexity of computing the Markov Blanket is $\calO(n^4)$. The sample complexity for the black-box queries is  $\calO(\max ( \frac{n k^3 \log^5 k}{\epsilon^2} , \frac{n k^3}{\epsilon^2} \log \frac{1}{\delta}\log k ))$  and the time complexity is $\calO(n k^4 \sqrt{k} \log k)$ (See Appendix \ref{proof_sample_time_obs_data} for details).
	
For synthetic experiments validating our theory, please See Appendix \ref{sec_experiments}.


\paragraph{Concluding Remarks.}
\label{sec_conclusion}
In this paper, we provide a novel method with theoretical guarantees to recover directed structure of a Bayesian network using black-box queries. We further improve our results when we have access to some observational data. We developed a theory for rank $2$ CPTs which can easily be extended to a more general rank $k$ CPTs. It would be interesting to see if we can provide similar results for a Bayesian network with low rank CPTs using pure observational or interventional data. 



\newpage
\appendix

\section*{Appendix}

\section{Detailed Proofs of Theorem and Lemmas}
\label{proof_detailed_proofs}
\subsection{Proof of Theorem \ref{thm_terminal_nodes_with_rank_2}}
\label{proof_thm_terminal_nodes_rank_2}

\emph{ \paragraph{Theorem \ref{thm_terminal_nodes_with_rank_2}}
   If nodes of a Bayesian network $\calB(G, \calP)$ have rank $2$ with respect to their parents $\pi_G(.)$ and $\calP$, then the Fourier coefficient $\hat f_t(B)$ for function $f_t$ defined by equation \eqref{eq_set_function_terminal_nodes} for any terminal node $t$ and a set $B \in 2^{\overline t}$ is given by:
   \begin{align}
   \hat f_t(B) = \begin{cases}
   Q_t(X_t = 1) + \frac{1}{2} \sum_{j \in \pi_G(t)} \big( Q_{tj}(X_t = 1, X_j = 0) + Q_{tj}(X_t = 1, X_j = 1)  \big), \quad B = \phi \\
   \frac{1}{2} \big( Q_{tj}(X_t = 1, X_j = 0) - Q_{tj}(X_t = 1, X_j = 1) \big), \quad B =\{j\}, \forall j \in \pi_G(t) \\
   0, \quad \text{Otherwise}
   \end{cases}
   \end{align}
}
\begin{proof}
	The Fourier transformation coefficients $\hat{f}_t$ can be calculated using the following formula:
	\begin{align}
	\label{eq_fourier_coeff_formula}
	\hat f_t(B) = 2^{-n+1} \sum_{A \in 2^{\overline t}} f_t(A) (-1)^{|A \cap B|}
	\end{align}
	
	We prove our claim by computing $\hat f_t (B)$ explicitly for various setting of $B \in 2^{\overline t}$.
	
	\paragraph{Case 1. $B = \phi $.}
	\begin{align*}
	\hat f_t(B) &= 2^{-n + 1} \sum_{A \in 2^{\overline t}} f_t(A) (-1)^{|A \cap B|} \\
	&= 2^{-n + 1} \sum_{A \in 2^{\overline t}} f_t(A), \text{ $\quad |A \cap B| = 0$ } \\
	&= 2^{-n + 1} \sum_{A \in 2^{\overline t}}  [Q_t(X_t = 1)	+ \sum_{j \in \pi_G(t)} Q_{tj}(X_t = 1, X_j = x_j^A]\\
	&= 2^{-n + 1} 2^{n - 1} Q_t(X_t = 1) + 2^{-n + 1} \sum_{A \in 2^{\overline t}}  \sum_{j \in \pi_G(t)} Q_{tj}(X_t = 1, X_j = x_j^A) \\
	&= Q_t(X_t = 1) + \frac{1}{2} \sum_{j \in \pi_G(t)} [Q_{tj}(X_t = 1, X_j = 0) + Q_{tj}(X_t = 1, X_j = 1)]
	\end{align*}
	
	\paragraph{Case 2. $B = \{l\}, l \in \pi_G(t) $.}
	\begin{align*}
	\hat f_t(B) &= 2^{-n+1} \sum_{A \in 2^{\overline t}} f_t(A) (-1)^{|A \cap B|} \\
	&= 2^{-n+1} [ - \sum_{A \in 2^{\overline t }, l \in A} f_t(A) + \sum_{A \in 2^{\overline t}, l \notin A} f_t(A) ] \\
	&= 2^{-n+1} [ - \sum_{A \in 2^{\overline t}, l \in A}  [Q_t(X_t = 1) + Q_{tl}(X_t = 1, X_l = 1) + \sum_{j \in \pi_G(t) - l} Q_{tj}(X_t = 1, X_j = x_j^A)] \\
	&+ \sum_{A \in 2^{\overline t}, l \notin A}  [Q_t(X_t = 1) + Q_{tl}(X_t = 1, X_l = 0) + \sum_{j \in \pi_G(t) - l} Q_{tj}(X_t = 1, X_j = x_j^A)]  ]\\
	&= \frac{1}{2} [ Q_{tl}(X_t = 1, X_l = 0) -  Q_{tl}(X_t = 1, X_l = 1) ]  \\
	\end{align*}
	
	\paragraph{Case 3. $B \subseteq \pi_G(t), |B| > 1 $.}
	\begin{align*}
	\hat f_t(B) &= 2^{-n+1} \sum_{A \in 2^{\overline t}} f_t(A) (-1)^{|A \cap B|} \\
	&= 2^{-n+1}   \sum_{A \in 2^{\overline t}} [ [Q_t(X_t = 1) + \sum_{j \in \pi_G(t) } Q_{tj}(X_t = 1, X_j = x_j^A)]  (-1)^{|A \cap B|}]\\
	&\text{Take an $l \in B \implies l \in \pi_G(t)$} \\
	&= 2^{-n+1} [ \sum_{A \in 2^{\overline t}, l \notin A}  [Q_t(X_t = 1) + Q_{tl}(X_t = 1, X_l = 0)  \\ &+\sum_{j \in \pi_G(t) - l} Q_{tj}(X_t = 1, X_j = x_j^A)] (-1)^{|A \cap B - l|}
	+  \sum_{A \in 2^{\overline t}, l \in A}  [Q_t(X_t = 1) + Q_{tl}(X_t = 1, X_l = 1) \\
	&+ \sum_{j \in \pi_G(t) - l} Q_{tj}(X_t = 1, X_j = x_j^A)] (-1)^{1 + |A \cap B - l|} ]\\
	&= 2^{-n+1} [ \sum_{A \in 2^{\overline t\setminus \{l\}}} [Q_{tl}(X_t = 1, X_l = 0) - Q_{tl}(X_t = 1, X_l = 1)] (-1)^{|A \cap B |}] \\
	&\text{Take $k \in B $} \\
	&= 2^{-n+1} [ \sum_{A  \in 2^{\overline t\setminus \{l\}}, k \in A} [Q_{tl}(X_t = 1, X_l = 0) - Q_{tl}(X_t = 1, X_l = 1)] (-1)^{1 + |A \cap B - k|}] \\
	&+ 2^{-n+1} [ \sum_{A  \in 2^{\overline t\setminus \{l\}}, k \notin A} [Q_{tl}(X_t = 1, X_l = 0) - Q_{tl}(X_t = 1, X_l = 1)] (-1)^{|A \cap B - k|}]\\
	&= 0
	\end{align*}
	
	\paragraph{Case 4. $ |B \cap \overline t-\pi_G(t)| \geq 1$}
	\begin{align*}
	\hat f_t(B) &= 2^{-n+1} \sum_{A \in 2^{\overline t}} f_t(A) (-1)^{|A \cap B|} \\
	&= 2^{-n+1} \sum_{A \in 2^{\overline t}} f_t(A \cap \pi_G(t)) (-1)^{|A \cap B|} \\
	&\text{Take $l \in B$ and $l \notin \pi_G(t)$}\\
	&= 2^{-n+1} [ \sum_{A \in 2^{\overline t}, l \notin A}  f_t(A \cap \pi_G(t)) (-1)^{|A \cap B - l|}  +  \sum_{A \in 2^{\overline t}, l \in A}  f_t(A \cap \pi_G(t)) (-1)^{1 + |A \cap B - l|} ]\\
	&= 0
	\end{align*}
	This proves our claim.
\end{proof}

\subsection{Proof of Theorem \ref{thm_nonterminal_nodes_with_rank_2}}
\label{proof_thm_nonterminal_nodes_with_rank_2}

\emph{\paragraph{Theorem \ref{thm_nonterminal_nodes_with_rank_2}} 
	If nodes of a Bayesian network $\calB(G, \calP)$ have rank $2$ with respect to their parents $\pi_G(.)$ and $\calP$, then the Fourier coefficient $\hat f_i(B)$ for function $f_i$ defined by equation \eqref{eq_set_function_nonterminal_nodes} for any non-terminal node $i$ and a set $B \in 2^{\overline i}$ is given by:
	\begin{align}
	\hat f_i(B) = \begin{cases}
	0, \quad | B\setminus \mb_G(i) | \geq 1  \\
	\frac{1}{2^{n - 1}} \sum_{A \in 2^{V - i}}  \frac{ g_i(A ) {\prod_{k \in \child_G(i)}} g_k(A) }{g_i(A ) {\prod_{k \in \child_G(i)}} g_k(A) +  g_i(A \cup \{i\}) {\prod_{k \in \child_G(i)}} g_k(A \cup \{i\}) } \psi_B(A) , \quad \text{otherwise}
	\end{cases}
	\end{align}
}
\begin{proof}
Note that for the case $| B - \mb_G(i) | = 0$, we simply replace terms in Equation \eqref{eq_set_function_nonterminal_nodes} with appropriate set functions. It can be simplified for various cases but we chose not to do it for clarity of representation. For the second case when $| B - \mb_G(i) | \geq 1$, $\exists s$ such that $s \in B$ and $s \notin \mb_G(i)$. Note that $f_i(A) = f_i(A \cap \mb_G(i))$. Take $A = A' \cup \{s\}$ and $s \notin A'$.
\begin{align*}
\hat f_i(B) &= 2^{-n+1} \sum_{A \in 2^{V - i}} f_i(A) (-1)^{|A \cap B|} \\
&= 2^{-n+1} (\sum_{A' \in 2^{V -\{i, s\}}} f_i(A') (-1)^{|A' \cap B| + 1} + \sum_{A' \in 2^{V - \{i, s\}}} f_i(A') (-1)^{|A' \cap B|})
&= 0
\end{align*}

\end{proof}

\newnote{
\subsection{Proof of Theorem \ref{thm_Stobbe}}
\label{proof_thm_Stobbe}
\emph{\paragraph{Theorem \ref{thm_Stobbe}}
Suppose $\bm{\hat g_i}$ is constructed by computing $\hat g_i(B_k)$ using $B_k$ from a fixed collection $ \rho_i$ as defined in Equation \eqref{eq_rho}. Furthermore, suppose $\bm{g_i}$ is computed by selecting $m_i$ sets $A_j$ uniformly at random from $2^{\overline i}$. We define the matrix $\mathcal{M}_i$ as in equation \eqref{eq_compressed_sensing_formulation}. Then there exist universal constants $C_1, C_2 > 0$ such that if, $m_i \geq \max( C_1 |\supp(\hat g_i)| \log^4 ( n + {{n - 1} \choose 2} ), C_2 |\supp(\hat g_i)| \log \frac{1}{\delta}  )$
and $\bm{\beta_i}$ is solved using equation \eqref{eq:opt prob}. Then with probability at least $1 - \delta$, we have $\| \bm{\beta_i} - \bm {\hat g_i} \|_2 \leq C_3 \frac{\epsilon}{\sqrt{m_i}}$ for some universal constant $C_3 > 0$.  If the minimum non-zero element of $ |\bm{\hat g_i}| $ is greater than $2 C_3  \frac{\epsilon}{\sqrt{m_i}} $ then $\bbeta_i$ recovers $\bm{\hat g_i}$ up to the signs. Furthermore, if Assumption \ref{assum_non_terminal_not_rank_2} is satisfied then $|\bbeta_i(B)| \leq C_3  \frac{\epsilon}{\sqrt{m_i}} , \forall B \in \rho_i, |B| = 2$ if and only if $i$ is a terminal node and  $\hat{\pi}(i) = \{B \mid |B| = 1, \ |\bbeta_i(B)| > C_3  \frac{\epsilon}{\sqrt{m_i}} \} $ correctly recovers the parents of the terminal node $i$, i.e., $\hat{\pi}(i) = \pi_G(i)$. Applying this recursively shows the correctness of Algorithm \ref{algo_getParents}. 
}
\begin{proof}
	First note that the rows of $\mathcal{M}_i$ are sampled uniformly at random from an orthonormal matrix with bounded entries. \cite{rauhut2010compressive} have proved that Restricted Isometry Property (RIP) holds for such matrices with high probability. Thus, we can invoke Theorem 1 from \cite{stobbe2012learning} which in turn follows the proof of Theorem 4.4 from \cite{rauhut2010compressive} to get the result that
	$\| \bm{\beta_i} - \bm {\hat g_i} \|_2 \leq C_3 \frac{\epsilon}{\sqrt{m_i}}$.
	
	Furthermore, $\| \bbeta_i - \bm{\hat g_i} \|_{\infty} \leq \| \bbeta_i - \bm{\hat g_i} \|_2 \leq C_3 \frac{\epsilon}{\sqrt{m_i}}   $. Thus if the minimum non-zero element of $ |\bm{\hat g_i}| $ is greater than $2 C_3  \frac{\epsilon}{\sqrt{m_i}} $ then $\bbeta_i$ recovers $\bm{\hat g_i}$ up to the signs.
	
	Adding to the above, the results from Theorem \ref{thm_terminal_nodes_with_rank_2} and Assumption \ref{assum_non_terminal_not_rank_2} ensure that $|\bbeta_i(B)| \leq C_3  \frac{\epsilon}{\sqrt{m_i}} , \forall B \in \rho_i, |B| = 2$ if and only if $i$ is a terminal node and $\hat{\pi}(i) = \pi_G(i)$.
\end{proof}
}

\subsection{Proof of Theorem \ref{thm_recover_mb}}
\label{proof_thm_recover_mb}
\emph{\paragraph{Theorem \ref{thm_recover_mb}}
	If there exists a probability distribution $\hat{P} \in \prob$ such that each node $i$ is rank $2$ with respect to $\mb_G(i)$ and $\hat{P}$, then the Markov blanket of a node $i$ can be recovered by solving the following system of equations:
	\begin{align}
	\begin{split}
	&\calP(X_i = 0, X_l=0) =  \tilde{Q}_i(X_i = 0) \calP(X_l=0) + \sum_{\substack{j \in -i \\ j \ne l}} \tilde{Q}_{ij}(X_i = 0, X_j = 0)  \calP( X_j = 0, X_l=0) \\
	&\hspace{1.3in}+ \tilde{Q}_{il}(X_i = 0, X_l=0) \calP(X_l = 0 ),\ \forall\ l=\{1,\dots,n \}, l \ne i \\
	&\calP(X_i = 0) =  \tilde{Q}_i(X_i = 0)  + \sum_{\substack{j \in -i \\ j \ne l}} \tilde{Q}_{ij}(X_i = 0, X_j = 0)  \calP( X_j = 0)
	\end{split}
	\end{align}
	which can be written in a more compact form:
	\begin{align}
	\overline \by &= \overline \bA \bq 
	\end{align}
	where $\overline \by \in \real^n$ and $\overline \bA \in \real^{n \times n}$ and $\bq \in \real^n$.
}
\begin{proof}
	If there exists a probability distribution $\hat{P} \in \prob$ such that each node $i$ is rank $2$ with respect to $\mb_G(i)$ and $\hat{P}$, then
	\begin{align}
	\label{eq_rank2_phat}	
	\hat{P}(X_i = x_i | X_{-i}) = Q_i(X_i = x_i) + \sum_{j \in -i } Q_{ij}(X_i = x_i, X_j)  
	\end{align}
	where $Q_{ij}(X_i=x_i, X_j) = 0$ if $j \notin \mb_G(i)$.
	
	For nodes $i, l \in \{1, \dots, n\}$ and $l \ne i$, consider the following:
	\begin{align}
	\label{eq_compute_Q}
	\begin{split}
	\hat{P} (X_i = x_i| X_l = x_l) = & \sum_{X_{-i - \{l\}}} \hat{P}(X_i = x_i, X_{-i - \{l\}} | X_l = x_l) \\
	=  &\sum_{X_{-i - \{l\}}} \hat{P}(X_i = x_i | X_{-i - \{l\}}, X_l = x_l)  \hat{P}( X_{-i - \{l\}} | X_l = x_l) \\
	= &\sum_{X_{-i - \{l\}}} \hat{P}(X_i = x_i | X_{-i})  \hat{P}( X_{-i - \{l\}} | X_l = x_l) \\
	&\text{Node $i$ is rank 2 with respect to $\hat{P}$ and $\mb_G(i)$} \\
	= &\sum_{X_{-i - \{l\}}} (Q_i(X_i = x_i) + \sum_{j \in -i} Q_{ij}(X_i = x_i, X_j))  \hat{P}( X_{-i - \{l\}} | X_l = x_l) \\
	= &Q_i(X_i = x_i) + \sum_{\substack{j \in -i \\ j \ne l}} \sum_{X_j} Q_{ij}(X_i = x_i, X_j)  \hat{P}( X_j | X_l = x_l) \\
	&+ Q_{il}(X_i = x_i, X_l = x_l) \\
	&\text{Now $\hat{P} \in \prob$}\\
	\calP(X_i = x_i| X_l = x_l) =& Q_i(X_i = x_i) + \sum_{\substack{j \in -i \\ j \ne l}} \sum_{X_j} Q_{ij}(X_i = x_i, X_j)  \calP( X_j | X_l=x_l) \\
	&+ Q_{il}(X_i = x_i, X_l = x_l) \\
	\calP(X_i = x_i, X_l = x_l) =& Q_i(X_i = x_i) \calP(X_l = x_l) + \sum_{\substack{j \in -i \\ j \ne l}} \sum_{X_j} Q_{ij}(X_i = x_i, X_j)  \calP( X_j, X_l = x_l) \\
	& + Q_{il}(X_i = x_i, X_l = x_l) \calP(X_l = x_l)
	\end{split}
	\end{align}
	
	We only focus on the case when $x_i = 0$ because that would be sufficient to determine the Markov Blanket for node $i$. Equation \eqref{eq_compute_Q} may not have a unique solution because for any pair of nodes $i, j$ if $Q_i(X_i = 0)$, $Q_{ij}(X_i = 0, X_j=0)$ and $Q_{ij}(X_i = 0, X_j=1)$ are part of a solution then there exists a solution with $Q_i(X_i = 0) + \epsilon$, $Q_{ij}(X_i = 0, X_j=0) - \epsilon$ and $Q_{ij}(X_i = 0, X_j=1) - \epsilon$. We focus on a particular solution where $\tilde{Q}_i(X_i = 0) = Q_i(X_i = 0) + \sum_{j \in \mb_G(i)} Q_{ij}(X_i=0, X_j = 1)$, $\tilde{Q}_{ij}(X_i = 0, X_j = 0) = Q_{ij}(X_i=0, X_j=0) - Q_{ij}(X_i=0, X_j=1)$ and thus equation \eqref{eq_compute_Q} becomes:
	\begin{align}
	\label{eq_compute_Q_tilde}
	\begin{split}
	\calP(X_i = 0, X_l=x_l) = & \tilde{Q}_i(X_i = 0) \calP(X_l=x_l) + \sum_{\substack{j \in -i \\ j \ne l}} \tilde{Q}_{ij}(X_i = 0, X_j = 0)  \calP( X_j = 0, X_l=x_l) \\
	&+ \tilde{Q}_{il}(X_i = 0, X_l=x_l) \calP(X_l = x_l ),\ \forall\ l=\{1,\dots,n \}, l \ne i, x_l \in \{0, 1\} 
	\end{split}
	\end{align}
	
	Equation \eqref{eq_compute_Q_tilde} can be written as a system of linear equations:
	\begin{align}
	\label{eq_compute_Q_tilde_matrix}
	\by &= \bA \bq 
	\end{align}
	where $\by \in \real^{2 n - 2}, \bA \in \real^{2n - 2 \times n}$ and $\bq \in \real^n$. We define $\bq$ as follows:
	\begin{align}
	\bq_j = \begin{cases} \tilde{Q}_{ij}(X_i=0, X_j=0), \quad \text{if }j < i \\ \tilde{Q}_{ij+1}(X_i=0, X_{j+1}=0), \quad \text{if } i \leq j \leq n -1 \\  \tilde{Q}_i(X_i=0), \quad \text{if }j = n \end{cases} \forall j \in \{1,\dots,n\}
	\end{align}
	The rows of $\by$ and $\bA$ are indexed by $l$ and $X_l$, i.e., 
	\begin{align}
	\begin{split}
	\by(l, X_l=x_l) &= \calP(X_i=0, X_l=x_l) \\
	\bA(l, X_l = x_l) &= \begin{cases} \calP(X_j = 0, X_l=x_l) , \quad \text{if }j < i \\
	\calP(X_{j+1} = 0, X_l=x_l) , \quad \text{if } i \leq j \leq n -1 \\
	\calP(X_l = x_l)  , \quad \text{if }j = n
	\end{cases}
	\end{split}
	\end{align}
	We take $ \calP(X_l = 0, X_l = 0) = \calP(X_l=0) $ and $ \calP(X_l = 0, X_l = 1) = 0 $.
	We can remove the linearly dependent rows from the above system of equations. For simplicity, let us assume that $i=n$. Then for $l = \{2,\dots, n-1\} $, $\by(1, X_1 = 0) + \by(1, X_1 = 1) - \by(l, X_l = 0) = [ \bA(1, X_1 = 0) + \bA(1, X_1 = 1) - \bA(l, X_l = 0)] \bq$ is equivalent to $  \by(l, X_l = 1) = \bA(l, X_l = 1) \bq$. Thus we can remove all the rows of $\by$ and $\bA$ indexed by $l, X_l = 1, \forall l = \{2,\dots, n-1\}$ and replace the last row of $\by$ and $\bA$ by $\by(1, X_1 = 0) + \by(1, X_1 = 1)$  and $\bA(1, X_1 = 0) + \bA(1, X_1 = 1)$ respectively. A similar argument can be presented for the case when $i \ne n$. 
\end{proof}

\subsection{Proof of Lemma \ref{lem_pso_sem_def_bar_A}}
\label{proof_lem_pso_sem_def_bar_A}
\emph{\paragraph{Lemma \ref{lem_pso_sem_def_bar_A}}
	The population matrix $\overline \bA$ as defined in equation \eqref{eq_compute_Q_tilde_matrix_compact} is a positive semidefinite matrix.
	}
\begin{proof}
	Here we carry out the proof for $i = n$. The same argument can be applied when $i \ne n$. Consider a random vector $\bz \in \real^n$ such that $z_j = \bm{1}[X_j = 0], \forall j=\{1,\dots, n-1\}$ and $z_n = 1$. Note that $\calP(X_i = 0) = \E[\bm{1}[X_i = 0]] = \E[\bm{1}[X_i = 0]^2]$ and $\calP(X_i = 0, X_j = 0) = \E[ \bm{1}(X_i = 0) \bm{1}(X_j = 0)], \forall i, j \in \{1, \dots, n\}$. Thus $\overline \bA = \E[\bz \bz^\T]$ which is a positive semidefinite matrix.
\end{proof}

\subsection{Proof of Lemma \ref{lem_only_recover_mb}}
\label{proof_lem_only_recover_mb}
\emph{\paragraph{Lemma \ref{lem_only_recover_mb}}
	If $\tilde{Q}_{ij}(\cdot,\cdot), \forall j \in \{1,\dots,n\}, j \ne i$ is computed by solving system of linear equations \eqref{eq_compute_Q_tilde_matrix_compact} and $\hat P \in \prob$ is faithful to $G$ then $ \tilde{Q}_{ij}(\cdot,\cdot) \ne 0, \forall j \in \{1,\dots, n\}, j \ne i$ if and only if $j \in \mb_G(i)$. 
}
\begin{proof}
	For the first part, suppose $\exists j \notin \mb_G(i)$ for which $\tilde{Q}_{ij}(\cdot,\cdot) \ne 0$, then expanding $\hat P(X_i | X_{-i}) = \tilde{Q}_i(\cdot) + \sum_{j=1, j \ne i}^n \tilde{Q}_{ij}(\cdot,\cdot)$, we see that $\hat P(X_i | X_{-i}) \ne \hat P(X_i | X_{\mb_G(i)})$ which violates the faithfulness assumption. For the reverse, suppose $\exists j \in \mb_G(i)$ for which $\tilde{Q}_{ij}(\cdot,\cdot) = 0$. This implies that $X_i$ and $X_j$ are independent given all the other nodes which again violates faithfulness.   
\end{proof}

\subsection{Proof of Lemma \ref{lem_observation_samples}}
\label{proof_lem_observation_samples}
\emph{\paragraph{Lemma \ref{lem_observation_samples}}
	 $N = \calO(\frac{\log n}{\epsilon^2})$ i.i.d observations are sufficient to measure elements of $\overline \bA$ and $\overline \by$, $\epsilon$ close to their true value. That is $| \overline \bA - \hat{\bA} |  \leq \epsilon$ and $|\overline \by-\hat\by| \leq \epsilon$, for some $\epsilon > 0$ with probability at least $1 - 2 \exp( \log ( {n \choose 2} + 3n )  - \frac{N \epsilon^2}{2})$ for some $\epsilon > 0$ where $\hat{\bA}$ and $\hat{\by}$ are the empirical measurements of $\overline {\bA}$ and $\overline {\by}$ respectively and $|\cdot - \cdot|$ denotes componentwise comparison for matrices.
}
\begin{proof}
For the observational data, we need to infer ${n \choose 2}$ probabilities of the form $P(X_i = 0, X_j = 0), \forall i, j \in \{1,\dots, n\}$, $n$ probabilities of the form $P(X_i = 0), \forall i=\{1,\dots, n\}$, $n$ probabilities each of the form $P(X_i = 0, X_1 = 1)$ and $P(X_i=0, X_2 = 1), \forall i=\{1,\dots,n\}$. Considering some ordering for $(X_i = x_i, X_j = x_j) = x_{ij}$. We consider $x_{ij} \leq x_{ij}'$ if $x_{ij}$ comes before $x_{ij}'$ in the ordering. Correspondingly, we can define the CDF $F_{ij}(x_{ij}) \triangleq \prob( (X_i, X_j) \leq x_{ij} )$. Now, we can apply Dvoretzky-Kiefer-Wolfowitz inequality\citep{dvoretzky1956asymptotic},
\begin{align}
\label{eq_DKW_inequalty_Fij}
\prob(\sup_{x_{ij}} |\hat F_{ij} (x_{ij}) -  F_{ij} (x_{ij}) | \geq \frac{\epsilon}{2} ) \leq 2 \exp(- \frac{N \epsilon^2}{2}), \forall \epsilon > 0
\end{align}
A similar equation can be written for the CDF of $P(X_i)$:
\begin{align}
\label{eq_DKW_inequalty_Fi}
\prob(\sup_{x_{i}} |\hat F_{i} (x_{i}) -  F_{i} (x_{i}) | \geq \frac{\epsilon}{2} ) \leq 2 \exp(- \frac{N \epsilon^2}{2}), \forall \epsilon > 0
\end{align}
where $N$ is number of i.i.d. samples. We compute actual probabilities by using the CDFs. For example:
\begin{align*}
\sup_{x_i } | \hat \prob(X_i = x_i) - \prob(X_i = x_i) | &= \sup_{x_i} | \hat F_{i} (x_i) - \hat F_{i} (x_i-1) - F_{i} (x_i) + F_{i} (x_i-1)  | \\
&\leq  \sup_{x_i} | \hat F_{i} (x_i) -  F_{i} (x_i)|  + \sup_{x_i } | \hat F_{i} (x_i - 1) - F_{i} (x_i-1)  | \\
&\leq \epsilon
\end{align*}
We need to ensure that this happens across all possible computations of probabilities. Thus taking a union bound,
\begin{align}
\label{eq_DKW_inequalty_union_bound_Fi_Fij}
&\prob((\exists X_i) \sup_{x} |\hat F_{i} (x_i) -  F_{i} (x_i) | \geq \frac{\epsilon}{2} \vee (\exists X_i, X_j) \sup_{x_{ij}} |\hat F_{ij} (x_{ij}) -  F_{ij} (x_{ij}) | \geq \frac{\epsilon}{2} ) \\
&\leq 4 \exp( \log ( {n \choose 2} + 3n )  - \frac{N \epsilon^2}{2}), \forall \epsilon > 0
\end{align}
\end{proof}

\subsection{Proof of Lemma \ref{lem_observ_nosiy_measurement}}
\label{proof_lem_observ_noisy_measurement}
\emph{\paragraph{Lemma \ref{lem_observ_nosiy_measurement}}
	Let $\hat\bA$ and $\hat\by$ be the empirical measurements of $\overline \bA$ and $\overline \by$ as defined in equation \eqref{eq_compute_Q_tilde_matrix_compact} respectively such that  $|\hat\bA - \overline \bA| \leq \epsilon$ and $|\hat\by - \overline \by| \leq \epsilon$ for some $\epsilon > 0$, where $|\cdot - \cdot|$ denotes componentwise comparison for matrices. Let $\hat\bq$ be the solution to the system of linear equations given by $\hat\by = \hat\bA \hat\bq $ and $\eta \kappa_{\infty}(\overline \bA) \leq 1$, then $\hat\bq$ recovers $\bq$ up to signs as long as $N = \calO(n)$ i.i.d. measurements are used to measure $\hat{\bA}$ and $ \frac{\max_i |\bq_i|}{\min_i |\bq_i|} \leq \frac{1 - \eta \kappa_{\infty}(\overline \bA)}{4 \eta \kappa_{\infty}(\overline \bA)}$, where $\kappa_{\infty}(\overline \bA) \triangleq \|\overline \bA\|_{\infty} \| \overline \bA^{-1} \|_{\infty}$ is the condition number of $\overline \bA$ and $\eta = \max ( \frac{n \epsilon}{ \sum_{j=1}^{n-1} \calP(X_j = 0 ) + 1 }, \frac{\epsilon}{\calP(X_n = 0)})$.
}
\begin{proof}
	Note that $\hat{\bA} \succ 0$ as long as $N = \calO(n)$ \cite{anderson1962introduction}. Here we carry out the proof for node $n$ but similar arguments hold for other nodes as well. First note that We denote $\Delta \bA \triangleq \hat\bA - \overline \bA$ and $\Delta \by \triangleq \hat\by - \overline \by$. First note that, $\| \overline \bA \|_{\infty} = \sum_{j=1}^{n-1} \calP(X_j = 0 ) + 1 $ and $\| \overline \by \|_{\infty} =  \calP(X_n = 0 ) $. Thus, $\| \Delta \bA \|_{\infty} \leq n \epsilon \leq \eta \|\overline \bA\|_{\infty} = \eta (\sum_{j=1}^{n-1} \calP(X_j = 0 ) + 1)$ and $\| \Delta \by \|_{\infty} \leq \epsilon \leq \eta \| \overline \by \|_{\infty} = \eta \calP(X_n = 0)$ for $\eta = \max ( \frac{n \epsilon}{ \sum_{j=1}^{n-1} \calP(X_j = 0 ) + 1 }, \frac{\epsilon}{\calP(X_n = 0)})$. Thus, we can invoke Theorem 2.2 from \cite{higham1994survey} and write,
	\begin{align}
	\begin{split}
	\frac{\| \hat\bq - \bq \|_{\infty}}{\| \bq \|_{\infty}} &\leq \frac{2 \eta \kappa_{\infty}(\overline \bA)}{1- \eta \kappa_{\infty}(\overline \bA)} \\
	\| \hat\bq - \bq \|_{\infty} &\leq \frac{2 \eta \kappa_{\infty}(\overline \bA)}{1- \eta \kappa_{\infty}(\overline \bA)} \| \bq \|_{\infty}
	\end{split}
	\end{align}
	It follows that if $ \frac{\max_i |\bq_i|}{\min_i |\bq_i|} \leq \frac{1 - \eta \kappa_{\infty}(\overline \bA)}{4 \eta \kappa_{\infty}(\overline \bA)}$ then we recover $\bq$ up to correct signs.
\end{proof}

\section{Sample and Time Complexity without access to any observational data}
\label{proof_sample_time_no_obs_data}
\paragraph{Sample Complexity.}
Using the Dvoretzky-Kiefer-Wolfowitz (DKW) inequality\citep{dvoretzky1956asymptotic} for each query independently and then taking the union bound across $m_i$ such queries, we get that each query is at max $\epsilon$ away from its true conditional probability with a probability of at least $1 - \sum_{i=1}^{m_i} 4\exp(-\frac{N_i \epsilon^2}{2})$. Let $N_{\min} \triangleq \min_{i=\{1,\dotsm m_i\}} N_i$ be the minimum number of sample we need across all query. Then we need $N_{\min} = \calO(\frac{\log m_i}{\epsilon^2})$ samples for each query to estimate probabilities of the form $\calP(X_i | X_{\overline {i}} = x_{\overline {i}})$, $\epsilon$ close to the true value with probability at least $1 -4 \exp(\log m_i - \frac{N_{\min} \epsilon^2}{2})$. The black-box outputs observational data for each of our queries independently and thus it needs to output a total of $\calO(\max ( \frac{n k^3 \log^4 n}{\epsilon^2} (\log k + \log \log n), \frac{n k^3}{\epsilon^2} \log \frac{1}{\delta}(\log k + \log \log n ))$ samples.

\paragraph{Time Complexity.}
\label{subsubsec_time_complexity}
Each optimization problem is solved using the logarithmic barrier method which takes $\calO(n^3 \sqrt{n} \log n)$ time. This needs to be repeated $\calO(nk)$ times. Thus, total time complexity is $\calO(n^4 k \sqrt{n} \log n)$.

\section{Sample and Time Complexity with access to some observational data}
\label{proof_sample_time_obs_data}
Regarding the black-box queries, we provide the same argument as Appendix \ref{proof_sample_time_no_obs_data} but for computing $\calP(X_i | X_{\mb_G(i)} = x_{\mb_G(i)})$, $\epsilon$ close to the true value with probability at least $1 - 4 \exp(\log m_i - \frac{N_{\min} \epsilon^2}{2}) $. We need to generate samples for each of our queries independently and thus need a total of $\calO(\max ( \frac{n k^3 \log^5 k}{\epsilon^2} , \frac{n k^3}{\epsilon^2} \log \frac{1}{\delta}\log k ))$  samples.

\paragraph{Time Complexity.}
\label{subsubsec_time_complexity_mixed_setting}
For the observational data, we are solving an optimization problem by computing inverse of a $\real^{n \times n}$ matrix and then multiplying it by a $\real^n$ vector. This can be done in $\calO(n^3)$ time. We repeat this process for each node, and thus it takes $\calO(n^4)$ time. All the inference can be done by only one traversal of the samples. Thus the total time complexity remains $\calO(n^4)$.

Regarding the black-box queries, each optimization problem is solved using the logarithmic barrier method which takes $\calO(k^3 \sqrt{k} \log k)$ time. This needs to be repeated $\calO(nk)$ times. Thus, the total time complexity is $\calO(n k^4 \sqrt{k} \log k)$.

\section{Synthetic Experiments}
\label{sec_experiments}

We conducted computational experiments on synthetic data to validate our results. In this section, we report the average performance across $5$ independently generated Bayesian networks. 

\begin{figure}[!ht]
	\centering
	\includegraphics[width=0.75\linewidth]{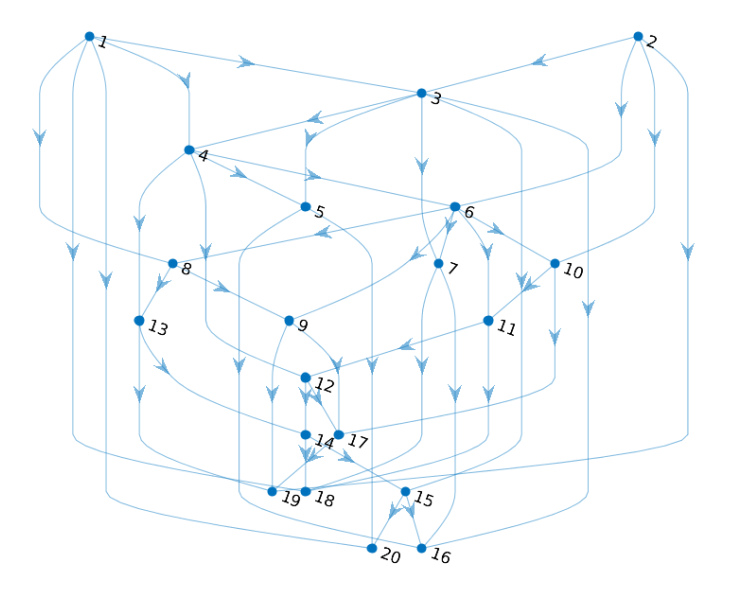}
	\caption{An example of synthetic Bayesian network generated on $n=20$ nodes}
	\label{fig:rep_graph}
\end{figure}

\paragraph{Generating Bayesian Networks.}
We generated $5$ synthetic Bayesian networks on $20$ nodes. We first chose a causal order for the nodes. We then generated CPTs for the nodes by making sure that each node's CPT is rank $2$ with respect to its parents. The parameters $Q_{ij}(\cdot, \cdot)$ as described in Equation \eqref{eq_rank_k_probability_distribution} were chosen uniformly at random from $[0, 1]$ while making sure that the resulting DAG is faithful. An example of a Bayesian network is shown in Figure \ref{fig:rep_graph}.

\paragraph{Black-box.}
We defined a black-box which can answer conditional probabilities queries $BB(i, A, x_A, N)$ to compute $\calP(X_i | X_A = x_A), \forall A \subseteq \{1,\dots,n\}$. The black-box outputs $N$ i.i.d. samples for $X_i$ given $X_A = x_A$.

\subsection{Recovering DAG without Access to any Observational Data.} 
For the first set of experiments, we did not have access to any observational data. Algorithm \ref{algo_getTerminalNodes} takes a Bayesian network on $S \subseteq \{1, \dots, n\}$ nodes and outputs terminal nodes $T \subseteq S$. The iterative use of Algorithm \ref{algo_getTerminalNodes} in Algorithm \ref{algo_getParents}, subsequently provides the exact DAG. We assume that the second node in the causal order does not have any parents. Following Theorem \ref{thm_Stobbe}, we submit $m_i = 10^C \max(k^2  \log^4 n', k^2  \log 1/\delta )$ queries for each node $i$ at each iteration where $k$ is the maximum number of nodes in Markov blanket, $n'$ is number of nodes in the Bayesian network at a specific iteration, i.e., $n' = |S|$ and $C$ is the control parameter. We fixed $k = 4$ and $\delta = 0.01$. The number of queries was capped at $300$ to ensure that we do not end up making too many queries. For each query, we only had access to $N = \calO(\frac{\log m_i}{\epsilon^2})$ samples from the black-box.

\paragraph{Results.}
We measured the performance of our method by measuring the Hamming distance between the true DAG and the recovered DAG. We also measured recall and precision for our method and then computed the $F1$ score to see their joint effect. The performance measures are defined formally as:
\begin{align*}
	\text{Hamming Distance} &= \sum_{i=1}^n (|\hat{\pi}(i)  \setminus \pi_G(i) | + |\pi_G(i)  \setminus \hat{\pi}(i)  |) \\
	\text{Precision} &= \frac{\sum_{i=1}^n |\hat{\pi}(i) \cap \pi_G(i)| }{\sum_{i=1}^n |\hat{\pi}(i)|} \\
	\text{Recall} &= \frac{\sum_{i=1}^n |\hat{\pi}(i) \cap \pi_G(i)| }{\sum_{i=1}^n |\pi_G(i)|} \\
	\text{F1 Score} &= \frac{2 \times \text{Precision} \times \text{Recall}}{\text{Precision} + \text{Recall}}
\end{align*}
where $\pi_G(i)$ is the set of true parents of node $i$ in true DAG $G$ and $\hat{\pi}(i)$ is the recovered set of parents of node $i$. Note that the recovery of a reversed edge is treated as a mistake. We show the average performance of our method across $5$ independently generated Bayesian networks. 

\begin{figure}[!ht]
	\centering
	\begin{subfigure}{.45\textwidth}
		\centering
		\includegraphics[width=\linewidth]{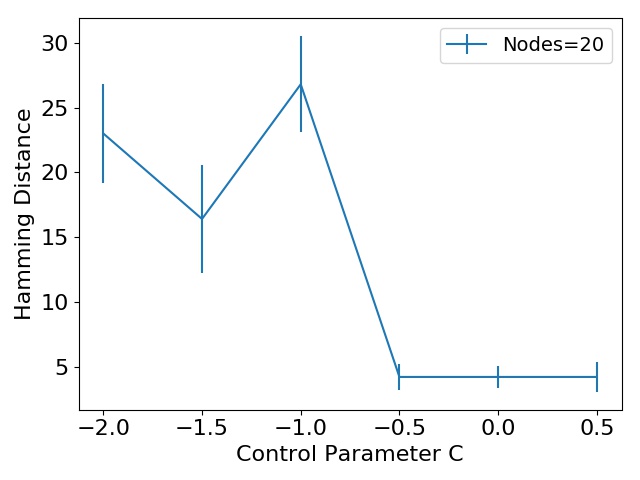}
		\caption{Hamming distance with control parameter $C$}
		\label{fig:hamming}
	\end{subfigure}%
	\begin{subfigure}{.45\textwidth}
		\centering
		\includegraphics[width=\linewidth]{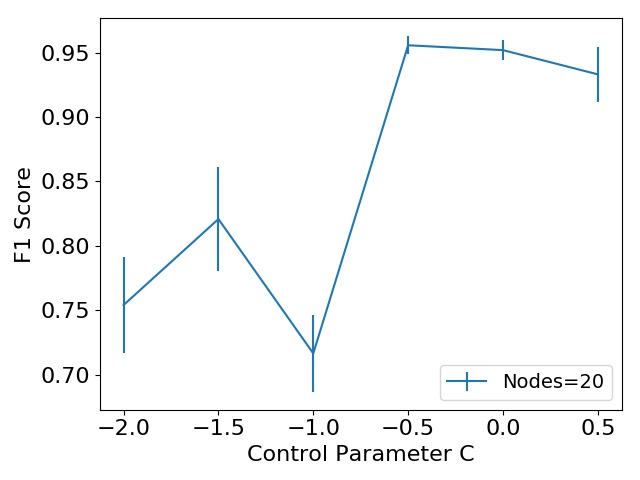}	
		\caption{$F1$ score with control parameter $C$}
		\label{fig:f1}
	\end{subfigure}
	\begin{subfigure}{.45\textwidth}
		\centering
		\includegraphics[width=\linewidth]{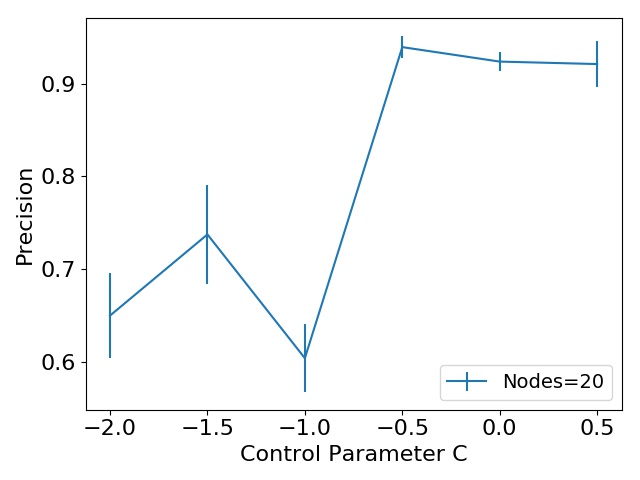}
		\caption{Precision with control parameter $C$}
		\label{fig:precision}
	\end{subfigure}%
	\begin{subfigure}{.45\textwidth}
		\centering
		\includegraphics[width=\linewidth]{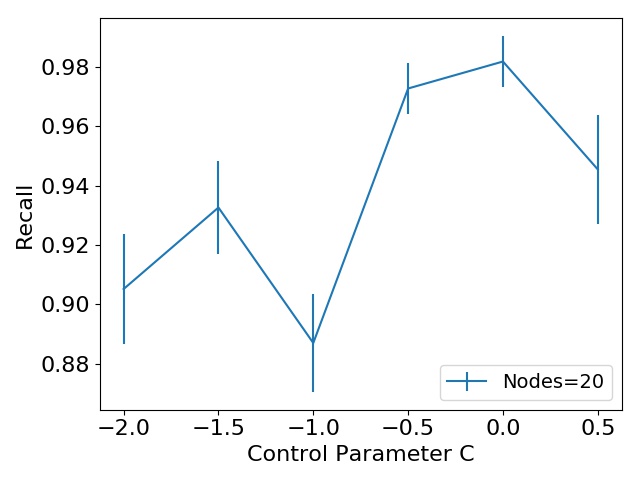}	
		\caption{Recall with control parameter $C$}
		\label{fig:recall}
	\end{subfigure}
	\caption{Regime without observational data. Plots of Hamming distance, $F1$ score, precision and recall versus the control parameter $C$ for Bayesian networks on $n = 20$ nodes with $m_i = 10^C \max(k^2  \log^4 n', k^2  \log 1/\delta )$ queries for each node $i$.}
	\label{fig:performances}
\end{figure}   

Observe that in Figure \ref{fig:hamming} the Hamming distance goes towards zero as we increase the number of samples, or equivalently, as we increase the control parameter $C$. Similarly, in Figure \ref{fig:precision}, \ref{fig:recall} both precision and recall (and $F1$ score as a result in Figure \ref{fig:f1}) go towards $1$ as we increase the control parameter $C$ in our experiments with a sharp transition around $C=-1$. This is consistent with our expected results from Theorem \ref{thm_Stobbe} and validates our theory. 

\subsection{Recovering Markov Blanket with Access to Some Observational Data.} 
For the second set of experiments, we had access to some observational data. Our method can be made more efficient by first computing the Markov blanket for a node and then applying Algorithm \ref{algo_getTerminalNodes} with queries of the form $f_i(A_j) = \calP(X_i \mid X_{A \cap \mb_G(i)} = x_{A \cap \mb_G(i)})$. Since, usually $|A| \gg |A \cap \mb_G(i)|$, this saves a lot of computational efforts and Black-box queries for our algorithm. Note that $n$ observations are necessary for Lemma \ref{lem_observ_nosiy_measurement} to work. Beyond this, from Lemma \ref{lem_observation_samples}, we only require $\calO(\frac{\log n}{\epsilon^2} )$ observational samples for recovering the Markov blankets of all the nodes. Thus, we conducted the experiments by generating $N = \max(10^C \frac{\log n}{\epsilon^2}, n)$ observational samples. The results of the experiments are provided below.

\paragraph{Results.}
As before, we measured performance of our method by measuring the Hamming distance between the true Markov blankets and the recovered ones. We also measured recall and precision for our method and then computed the $F_1$ score to see their joint effect. The performance measures are defined slightly differently as the recovery is with respect to the Markov blankets. 
\begin{align*}
\text{Hamming Distance} &= \sum_{i=1}^n (|\hat{\mb}(i)  \setminus \mb_G(i) | + |\mb_G(i)  \setminus \hat{\mb}(i)  |) \\
\text{Precision} &= \frac{\sum_{i=1}^n |\hat{\mb}(i) \cap \mb_G(i)| }{\sum_{i=1}^n |\hat{\mb}(i)|} \\
\text{Recall} &= \frac{\sum_{i=1}^n |\hat{\mb}(i) \cap \mb_G(i)| }{\sum_{i=1}^n |\mb_G(i)|} \\
\text{F1 Score} &= \frac{2 \times \text{Precision} \times \text{Recall}}{\text{Precision} + \text{Recall}}
\end{align*}
where $\mb_G(i)$ is the set of nodes in the Markov blanket of node $i$ in true DAG $G$ and $\hat{\mb}(i)$ is the recovered set of nodes in the Markov blanket of node $i$. Below we provide average performance of our method across $5$ independently generated Bayesian networks. 

\begin{figure}[!ht]
	\centering
	\begin{subfigure}{.45\textwidth}
		\centering
		\includegraphics[width=\linewidth]{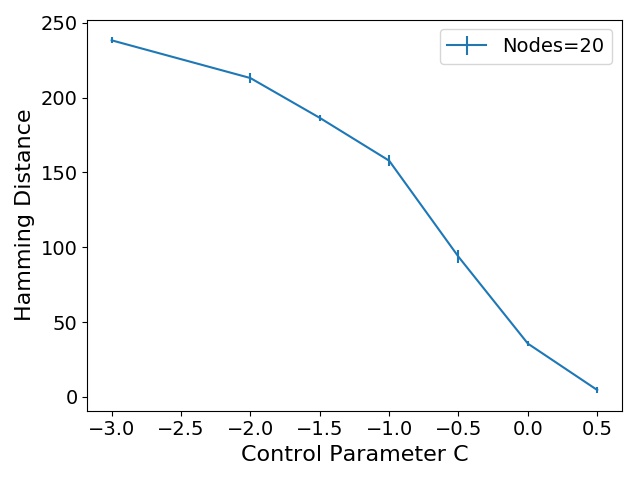}
		\caption{Hamming distance of Markov blanket recovery with control parameter $C$}
		\label{fig:hammingmb}
	\end{subfigure}%
	\begin{subfigure}{.45\textwidth}
		\centering
		\includegraphics[width=\linewidth]{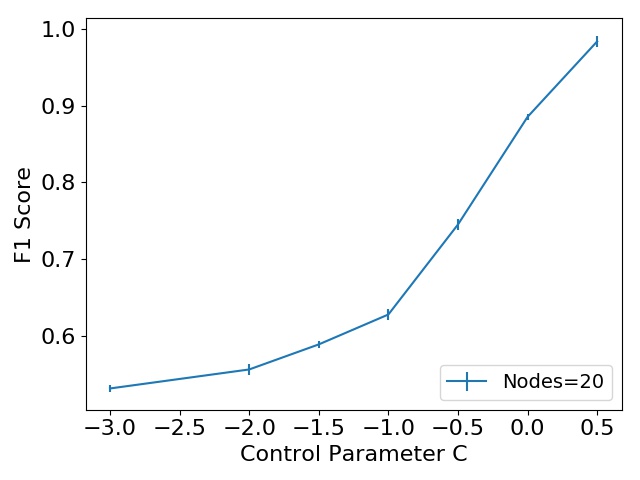}	
		\caption{$F1$ score of Markov blanket recovery with control parameter $C$}
		\label{fig:f1mb}
	\end{subfigure}
	\begin{subfigure}{.45\textwidth}
		\centering
		\includegraphics[width=\linewidth]{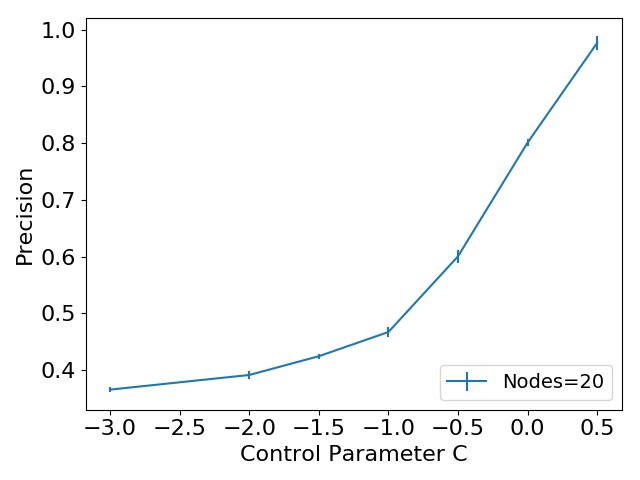}
		\caption{Precision of Markov blanket recovery with control parameter $C$}
		\label{fig:precisionmb}
	\end{subfigure}%
	\begin{subfigure}{.45\textwidth}
		\centering
		\includegraphics[width=\linewidth]{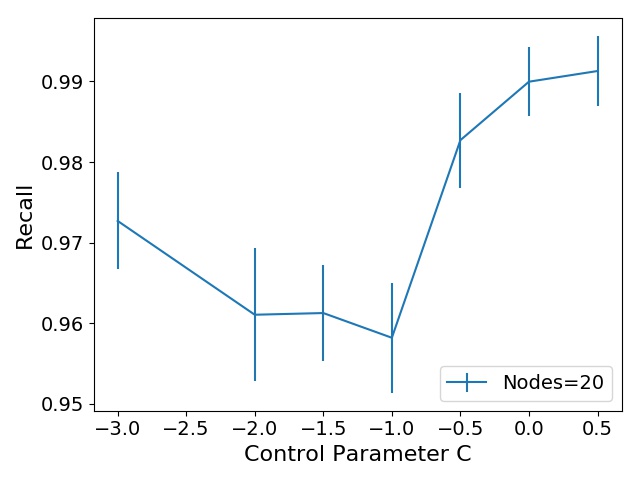}
		\caption{Recall of Markov blanket recovery with control parameter $C$}
		\label{fig:recallmb}
	\end{subfigure}
	\caption{Regime with observational data. Plots of Hamming distance, $F1$ score, precision and recall for Markov blanket recovery versus the control parameter $C$ for Bayesian networks on $n = 20$ nodes with $N = \max(10^C \frac{\log n}{\epsilon^2}, n) )$ observational samples}
	\label{fig:f1hammingmb}
\end{figure}   

We see in Figure \ref{fig:hammingmb} that the Hamming distance of Markov blanket recovery goes to zero as we increase number of observational samples, or equivalently, as we increase the control parameter $C$. Similarly, precision and recall of Markov blanket recovery in Figure \ref{fig:precisionmb}, \ref{fig:recallmb} approach  $1$ as number of observational samples increase. This validates our theory. Another interesting observation is that recall is very close to $1$ even for a small number of observational samples. This is good for our method as it would still work when recovering any set $S$ such that $\mb_G(i) \subseteq S$. The sample and time complexities are improved depending on the size of $S$ (the best result is achieved when $S = \mb_G(i)$). 

After we recovered the Markov blanket, we executed our Algorithm \ref{algo_getTerminalNodes} with $f_i(A_j) = \calP(X_i \mid X_{A \cap \mb_G(i)} = x_{A \cap \mb_G(i)})$. We then obtained similar results as in the no-observational-data regime, but with smaller number of samples and less computation.
 

\begin{thebibliography}{}
		\bibitem[Anderson, 1962]{anderson1962introduction}
		Anderson, T.~W. (1962).
		\newblock {An Introduction to Multivariate Statistical Analysis}.
		\newblock Technical report, Wiley New York.
		
		\bibitem[Bello and Honorio, 2018]{bello2018computationally}
		Bello, K. and Honorio, J. (2018).
		\newblock {Computationally and Statistically Efficient Learning of Causal Bayes
			Nets Using Path Queries}.
		\newblock In {\em Advances in Neural Information Processing Systems}, pages
		10931--10941.
		
		\bibitem[Cheng et~al., 2002]{cheng2002learning}
		Cheng, J., Greiner, R., Kelly, J., Bell, D., and Liu, W. (2002).
		\newblock {Learning Bayesian Networks From Data: An Information-Theory Based
			Approach}.
		\newblock {\em Artificial intelligence}, 137(1-2):43--90.
		
		\bibitem[Cussens, 2008]{cussens2012bayesian}
		Cussens, J. (2008).
		\newblock {Bayesian Network Learning by Compiling to Weighted MAX-SAT}.
		\newblock {\em Uncertainty in Artificial Intelligence}.
		
		\bibitem[D., 1996]{chickering1996learning}
		D., C. (1996).
		\newblock {Learning Bayesian Networks Is NP-Complete}.
		\newblock {\em Learning from Data}, pages 121--130.
		
		\bibitem[Dvoretzky et~al., 1956]{dvoretzky1956asymptotic}
		Dvoretzky, A., Kiefer, J., and Wolfowitz, J. (1956).
		\newblock {Asymptotic Minimax Character of the Sample Distribution Function and
			of the Classical Multinomial Estimator}.
		\newblock {\em The Annals of Mathematical Statistics}, pages 642--669.
		
		\bibitem[Eaton and Murphy, 2007]{eaton2007exact}
		Eaton, D. and Murphy, K. (2007).
		\newblock {Exact Bayesian Structure Learning From Uncertain Interventions}.
		\newblock {\em Artificial Intelligence and Statistics}, pages 107--114.
		
		\bibitem[Friedman et~al., 1999]{friedman1999learning}
		Friedman, N., Nachman, I., and Pe{\'e}r, D. (1999).
		\newblock {Learning Bayesian Network Structure From Massive Datasets: The
			Sparse Candidate Algorithm}.
		\newblock In {\em Proceedings of the Fifteenth conference on Uncertainty in
			artificial intelligence}, pages 206--215. Morgan Kaufmann Publishers Inc.
		
		\bibitem[Hausar and B{\"u}hlmann, 2012]{hausar2012optimal}
		Hausar, A. and B{\"u}hlmann, P. (2012).
		\newblock {Two Optimal Strategies for Active Learning of Causal Models From
			Interventions}.
		\newblock {\em Proceedings of the 6th European Workshop on Probabilistic
			Graphical Models}.
		
		\bibitem[He and Geng, 2008]{he2008active}
		He, Y. and Geng, Z. (2008).
		\newblock {Active Learning of Causal Networks With Intervention Experiments and
			Optimal Designs}.
		\newblock {\em Journal of Machine Learning Research}.
		
		\bibitem[Higham, 1994]{higham1994survey}
		Higham, N.~J. (1994).
		\newblock {\em {A Survey of Componentwise Perturbation Theory}}, volume~48.
		\newblock American Mathematical Society.
		
		\bibitem[Jaakkola et~al., 2010]{jaakkola2010learning}
		Jaakkola, T., Sontag, D., Globerson, A., and Meila, M. (2010).
		\newblock {Learning Bayesian Network Structure Using LP Relaxations}.
		\newblock In {\em Proceedings of the Thirteenth International Conference on
			Artificial Intelligence and Statistics}, pages 358--365.
		
		\bibitem[Kocaoglu et~al., 2017]{kocaoglu2017experimental}
		Kocaoglu, M., Shanmugam, K., and Bareinboim, E. (2017).
		\newblock {Experimental Design for Learning Causal Graphs With Latent
			Variables}.
		\newblock In {\em Advances in Neural Information Processing Systems}, pages
		7018--7028.
		
		\bibitem[Koivisto and Sood, 2004]{koivisto2004exact}
		Koivisto, M. and Sood, K. (2004).
		\newblock {Exact Bayesian Structure Discovery in Bayesian Networks}.
		\newblock {\em Journal of Machine Learning Research}, 5(May):549--573.
		
		\bibitem[Margaritis and Thrun, 2000]{margaritis2000bayesian}
		Margaritis, D. and Thrun, S. (2000).
		\newblock {Bayesian Network Induction via Local Neighborhoods}.
		\newblock In {\em Advances in neural information processing systems}, pages
		505--511.
		
		\bibitem[Moore and Wong, 2003]{moore2003optimal}
		Moore, A. and Wong, W.-K. (2003).
		\newblock {Optimal Reinsertion: A New Search Operator for Accelerated and More
			Accurate Bayesian Network Structure Learning}.
		\newblock In {\em International Conference on Machine Learning}, volume~3,
		pages 552--559.
		
		\bibitem[Murphy, 2001]{murphy2001active}
		Murphy, K.~P. (2001).
		\newblock {Active Learning of Causal Bayes Net Structure}.
		\newblock {\em Technical Report}.
		
		\bibitem[Rauhut, 2010]{rauhut2010compressive}
		Rauhut, H. (2010).
		\newblock {Compressive Sensing and Structured Random Matrices}.
		\newblock {\em Theoretical foundations and numerical methods for sparse
			recovery}, 9:1--92.
		
		\bibitem[Silander and Myllym{\"a}ki, 2006]{silander2012simple}
		Silander, T. and Myllym{\"a}ki, P. (2006).
		\newblock {A Simple Approach for Finding the Globally Optimal Bayesian Network
			Structure}.
		\newblock In {\em Uncertainty in Artificial Intelligence}, pages 445--452.
		
		\bibitem[Spirtes et~al., 2000]{spirtes2000causation}
		Spirtes, P., Glymour, C.~N., and Scheines, R. (2000).
		\newblock {\em {Causation, Prediction, and Search}}.
		\newblock MIT press.
		
		\bibitem[Stobbe and Krause, 2012]{stobbe2012learning}
		Stobbe, P. and Krause, A. (2012).
		\newblock {Learning Fourier Sparse Set Functions}.
		\newblock In {\em Artificial Intelligence and Statistics}, pages 1125--1133.
		
		\bibitem[Tong and Koller, 2001]{tong2001active}
		Tong, S. and Koller, D. (2001).
		\newblock {Active Learning for Structure in Bayesian Networks}.
		\newblock {\em International Join Conference on Artificial Intelligence},
		17:863--869.
		
		\bibitem[Triantafillou and Tsamardinos, 2015]{triantafillou2015constraint}
		Triantafillou, S. and Tsamardinos, I. (2015).
		\newblock {Constraint-Based Causal Discovery From Multiple Interventions Over
			Overlapping Variable Sets}.
		\newblock {\em Journal of Machine Learning Research}.
		
		\bibitem[Tsamardinos et~al., 2006]{tsamardinos2006max}
		Tsamardinos, I., Brown, L.~E., and Aliferis, C.~F. (2006).
		\newblock {The Max-Min Hill-Climbing Bayesian Network Structure Learning
			Algorithm}.
		\newblock {\em Machine learning}, 65(1):31--78.
		
		\bibitem[Xie and Geng, 2008]{xie2008recursive}
		Xie, X. and Geng, Z. (2008).
		\newblock {A Recursive Method for Structural Learning of Directed Acyclic
			Graphs}.
		\newblock {\em Journal of Machine Learning Research}, 9(Mar):459--483.
		
		\bibitem[Yehezkel and Lerner, 2005]{yehezkel2005recursive}
		Yehezkel, R. and Lerner, B. (2005).
		\newblock {Recursive Autonomy Identification for Bayesian Network Structure
			Learning}.
		\newblock In {\em Artificial Intelligence and Statistics}, pages 429--436.
		Citeseer.
\end{thebibliography}
\end{document}